\documentclass[lettersize,journal]{IEEEtran}
\usepackage{amsmath,amsfonts}
\usepackage{array}
\usepackage[caption=false,font=normalsize,labelfont=sf,textfont=sf]{subfig}
\usepackage{textcomp}
\usepackage{stfloats}
\usepackage{url}
\usepackage{verbatim}
\usepackage{graphicx}
\usepackage{cite}
\usepackage{color}
\usepackage{hyperref}
\usepackage{xcolor}
\usepackage[linesnumbered,ruled,vlined]{algorithm2e}
\usepackage{subfig} 
\usepackage{subfloat}
\usepackage{caption}
\usepackage{multirow}
\usepackage{booktabs}
\usepackage{amsmath}
\usepackage{cleveref}

\crefname{table}{Table}{Tables}
\crefname{figure}{Fig.}{Fig.}
\crefname{equation}{Eq.}{Eq.}
\crefname{section}{section}{section}
\crefname{algorithm}{Algorithm}{Algorithm}

\usepackage{amsmath}

\newcolumntype{P}[1]{>{\centering\arraybackslash}p{#1}}
\newcolumntype{C}[1]{>{\centering\arraybackslash}m{#1}}

\newcommand*{\rom}[1]{\expandafter\romannumeral #1}

\def \textHT [#1]{\color{red}$\mathbf{#1}$\color{black}}

\def \textLT
[#1]{\color{blue}$\mathbf{#1}$\color{black}}

\SetCommentSty{mycommfont}
\SetKwInput{KwInput}{Input}                
\SetKwInput{KwOutput}{Output}              
\let\oldnl\nl
\newcommand{\nonl}{\renewcommand{\nl}{\let\nl\oldnl}}

\DeclareMathOperator*{\argmin}{arg\,min}

\begin{document}
\title{BioSLAM: A Bio-inspired Lifelong Memory System for General Place Recognition}
\author{Peng Yin\textsuperscript{1,*},~\IEEEmembership{Member,~IEEE},
        Abulikemu Abuduweili\textsuperscript{1,*},
        Shiqi Zhao\textsuperscript{2} \\ 
        Changliu Liu\textsuperscript{1},~\IEEEmembership{Member,~IEEE}
        and Sebastian Scherer\textsuperscript{1},~\IEEEmembership{Senior Member,~IEEE}
\thanks{Peng Yin, Abulikemu Abuduweili, Changliu Liu, and Sebastian Scherer are with the Robotics Institute, Carnegie Mellon University, Pittsburgh, PA 15213, USA. {(pyin2, abulikea, cliu6, basti)@andrew.cmu.edu}.}
\thanks{Shiqi Zhao is with the University of California San Diego, La Jolla, CA 92093, USA. {(s2zhao@eng.ucsd.edu)}.}
\thanks{\textsuperscript{*}Authors Peng Yin and Abulikemu Abuduweili contributed equally.
}
\thanks{Corresponding author: Peng Yin (pyin2@andrew.cmu.edu)}
}

\markboth{IEEE Transactions on Robotics (T-RO). Preprint Version. August 2022}
{Yin \MakeLowercase{\textit{et al.}}: BioSLAM: A Bio-inspired Lifelong Memory System for General Place Recognition}

\maketitle
\begin{abstract}
We present BioSLAM, a lifelong SLAM framework for learning various new appearances incrementally and maintaining accurate place recognition for previously visited areas.
Unlike humans, artificial neural networks suffer from catastrophic forgetting and may forget the previously visited areas when trained with new arrivals.
For humans, researchers discover that there exists a memory replay mechanism in the brain to keep the neuron active for previous events.
Inspired by this discovery, BioSLAM designs a gated generative replay to control the robot's learning behavior based on the feedback rewards.
Specifically, BioSLAM provides a novel dual-memory mechanism for maintenance: 1) a dynamic memory to efficiently learn new observations and 2) a static memory to balance new-old knowledge.
When combined with a visual-/LiDAR- based SLAM system, the complete processing pipeline can help the agent incrementally update the place recognition ability, robust to the increasing complexity of long-term place recognition.

We demonstrate BioSLAM in two incremental SLAM scenarios.
In the first scenario, a LiDAR-based agent continuously travels through a city-scale environment with a 120km trajectory and encounters different types of 3D geometries (open streets, residential areas, commercial buildings).
We show that BioSLAM can incrementally update the agent's place recognition ability
and outperforming the state-of-the-art incremental approach, Generative Replay, by 24\%.
In the second scenario, a LiDAR-vision-based agent repeatedly travels through a campus-scale area on a 4.5km trajectory. 
BioSLAM can guarantee the place recognition accuracy to outperform 15\% over the state-of-the-art approaches under different appearances.
To our knowledge, BioSLAM is the first memory-enhanced lifelong SLAM system to help incremental place recognition in long-term navigation tasks.
\end{abstract}


\begingroup
\let\clearpage\relax
    \begin{figure}[t]
    \begin{center}
    \includegraphics[width=\linewidth]{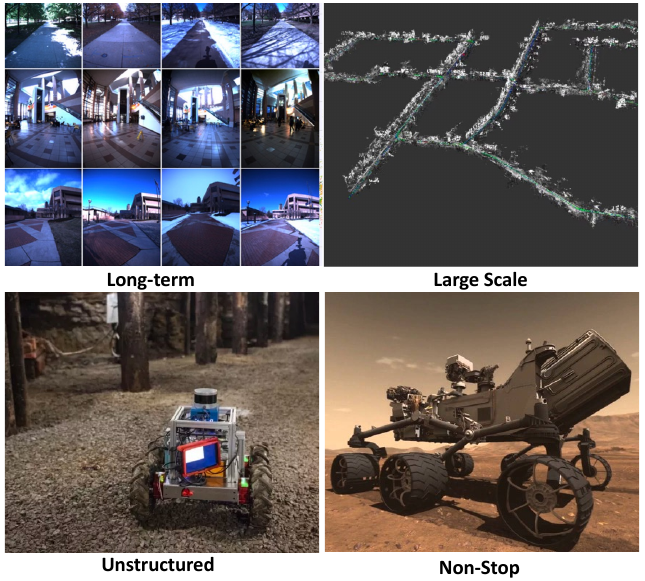}
    \end{center}
    \vspace{-5pt}
    \caption{\textbf{Challenges in Real-world Robotic Localization.}
    For real-world field applications, robotic localization usually encounters the following challenges: 1) changing appearance under long-term environmental variants, 2) diverse geometric differences under large-scale areas, 3) mixture structure/unstructured environments, and 4) non-stop restriction for long-term autonomy.}
    \label{fig:idea}
\end{figure}

\section{Introduction}
\label{sec:introduction}
\IEEEPARstart{A}{n} essential capability for long-term robotics autonomy in the open world without human assistance is life-long Simultaneous Localization and Mapping (SLAM)~\cite{Survey:SLAM}.
In the context of lifelong SLAM, the system needs to consider work in long-term operation in large-scale environments and diverse environmental conditions, as depicted in \cref{fig:idea}.
Current SLAM methods are mainly conducted under single-type environments, where the environmental conditions (such as illuminations, weather, seasons, etc.) are consistent, and these environments are mostly static.
Recent works attempt to relax the \textit{single-type} assumption to accommodate diverse environments by leveraging domain adaptation techniques into model learning with deep neural networks.
However, the learned place descriptors under new scenarios can affect the localization accuracy of previous scenarios, an effect known as ``catastrophic forgetting''.

In real-world long-term navigation, the robot may encounter complicated 3D environments, such as campus areas, open streets, residential blocks, commercial buildings, etc., and each place has its unique patterns in place recognition.
The robot platform can't collect datasets under all scenarios at once and train the localization module in a supervised manner.
A naive solution for incremental observations is to source additional data for model adaptation with a new scenario; however, this  adaptation is not feasible when the goal is to ensure the uninterrupted and long-term operation of the robot, since it causes catastrophic forgetting of previous knowledge.
Moreover, changes in environments can be sudden, e.g., rapid illumination and weather changes, while it may take too long for traditional learning-based approaches to react to the changes.  
As depicted in \cref{fig:idea}, the main challenges for lifelong place recognition include:

\begin{itemize}
    \item \textit{Various environmental conditions:} the appearances of the same area under different environmental conditions will be represented with different patterns. 
    \item \textit{Diverse scenarios:} the robot platform will encounter different 3D environments in large-scale navigation tasks, and most areas are a combination of different types.
    \item \textit{Non-stop training:} for long-term autonomy challenges, the robot will accumulate new datasets, and model fine-tuning is usually required to improve localization performance for new scenarios.
\end{itemize}
With the above challenges, traditional SLAM methods mainly work for short-term navigation tasks and can hardly deal with long-term data association.
The lack of domain adaptation in existing methods has become a major hurdle to achieving long-term robotic autonomy because robots will encounter boundless new scenarios in real applications.
For most place recognition methods~\cite{detone2018superpoint, latif2018addressing}, the addressed domain adaptation only considers unidirectional knowledge transfer from a single domain to another fixed domain, which cannot be generalized to open world situations, where new environments that the robot can encounter are infinite and previously known environments can be visited under diverse conditions.

In this work, we propose a lifelong localization system, BioSLAM, which defines a lifelong SLAM framework that can continuously adapt to new environments without sacrificing performance in previously seen environments.
In our previous work~\cite{i3dloc}, we notice that cross-domain appearance differences will significantly affect the localization performance;
the localization module encounters the catastrophic forgetting problem, where it is only robust to the most recently trained scenarios. 
In contrast, humans and animals do not suffer from catastrophic forgetting, and short-term and long-term memory mechanisms exist within the hippocampus~\cite{Hip:Place_cell} and the front lobe of the brain~\cite{frontal_lobe}, which plays the main role in lifelong knowledge updating.
Recently, new evidence from fMRI studies in humans~\cite{hippo_memory} finds that the hippocampus may `act as a librarian to retrieve the cortical books of memory', i.e., the hippocampus can index the memories for fast retrievals.
Inspired by the biological mechanism, we design two memory zones for BioSLAM, namely static memory zones (SMZ) for historical memory encoding with low frequency and dynamic memory zone (DMZ) for quickly memory reply, and propose a dual-memory selection mechanism to balance the short-term adaptation for new observations and long-term memory retention for historical knowledge.
Specifically, BioSLAM also develops a sleeping cycle for memory consolidation within SMZ, which is also inspired by a similar mechanism in the hippocampus~\cite{klinzing2019mechanisms}.
Based on the above mechanism, BioSLAM has the ability to achieve long-term place recognition.


The evaluation methods~\cite{VPR_Bench} for traditional place recognition using supervised learning approaches do not apply to lifelong systems. The performance of lifelong systems is reflected by the adaptation capability with respect to new observations and the long-term memory retention of previously visited areas.
In this work, we formulate two metrics, namely adaptation efficiency (AE) analysis and retention ability (RA) analysis, and perform extensive evaluation using two long-term datasets: 1) \textit{City Dataset}, which is focused on changing geometric patterns, and 2) \textit{Campus Dataset}, which is focused on changing illumination patterns.
The major contributions of this paper are as follows:
\begin{itemize}
    \item BioSLAM provides a systematic framework to learn about ever-changing environments without interruption.
    Using this framework, we enable the incremental place feature learning in the long-term autonomy.
    \item Within BioSLAM, we develop a dual-memory module, which includes 1) a dynamic memory zone with high-frequency updates for fast adaptation of new patterns and 2) a static memory zone with low-frequency updates for long-term memory retention.
    \item BioSLAM can perform non-stop online learning for new environments and provide lifelong re-localization ability for previously visited areas even under changing environmental conditions.
    Furtherly, the module design of BioSLAM makes it possible to be combined with arbitrary place descriptor learning modules.
    \item We developed extensive lifelong localization datasets and relative metrics to evaluate the lifelong localization performance
    and demonstrate a detailed analysis of adaptation efficiency and long-term memory retention ability.
\end{itemize}

In the rest of the paper, we will introduce the related works for place recognition and lifelong incremental learning in \cref{sec:related_works}.
\Cref{sec:system_oveview} gives the structural overview of BioSLAM. 
\Cref{sec:GPFL} and \cref{sec:BiLM} explain the details of the general place feature learning and bio-inspired lifelong memory, respectively.
The experiment setup and qualitative/quantitative analysis are given in \cref{sec:exp_setup} and \cref{sec:exp_analysis}.

    \section{Related Works}
\label{sec:related_works}
There are two important modules in lifelong localization: 1) place recognition and 2) lifelong learning.
Place recognition (PR) or Loop closure detection (LCD) has been studied for decades, as stated in~\cite{VPR:SURVEY, deep_pr_survey, deep_vpr_survey}, which mainly serves as the data association for large-scale re-localization and map optimization in SLAM tasks.
Lifelong learning, also known as continual, incremental, or sequential learning, aims at incrementally building up knowledge from a sequential data stream~\cite{lifelong_survey,CLRobot}, which is essential for long-term localization where robots will encounter many infinite environments.
In the following subsections, we will mainly introduce the related works in visual/LiDAR place recognition and recent lifelong learning works from a robotics perspective.

\subsection{Long-term Place Recognition}
\label{sec:related_pr}
Place recognition targets identifying the exact areas under different perspectives and environmental conditions~\cite{VPR:SURVEY}.
There are mainly addressed with two types of approaches, namely visual-based and LiDAR-based place recognition.

For visual place recognition, the visual inputs are usually affected by illuminations and viewpoints. 
The traditional geometry descriptors (e.g., scale-invariant feature transform (SIFT)~\cite{FEATURE:SIFT} and oriented FAST and rotated BRIEF (ORB)~\cite{feature:orb}) are widely used in visual place recognition because of their invariant properties to scale, orientation and illumination changes.
Based on these handcrafted features, FAB-MAP~\cite{fabmap} build a Bag-of-visual-words (BoW) architecture to achieve large-scale visual re-localization.
iBoW-LCD~\cite{ibow_lcd} uses an incremental BoW scheme based on binary descriptors to retrieve matched images more efficiently.
An~\textit{et al.} introduces FILD++~\cite{bow_fast}, an incremental loop closure detection approach via constructing a hierarchical small‐world graph.
With the booming of deep learning, new convolutional neural network (CNNs) features, such VGG~\cite{Feature:VGG}, ResNet~\cite{resnet}, Transformer~\cite{transformer}, provide significent improvements in feature/semantic extraction.
NetVLAD~\cite{NetVLAD} combined the CNN features and an differentiable VLAD~\cite{VLAD} layer to enable deep learning for visual place recognition; and based on~\cite{NetVLAD}, recent deep learning approaches~\cite{regionVLAD,2019conditionVLAD,hui2021pyramid} further improve the recognition accuracy by combining with different networks.


For LiDAR-based place recognition, LiDAR inputs will not be affected by environmental conditions, such as illumination, weather, and season differences.
In non-learning based 3D localization, M2DP~\cite{m2dp} and Scan-context~\cite{scancontext} utilize the histogram of LiDAR projection to achieve long-term 3D re-localization.
With the developments of 3D deep feature extraction, recent learning-based approaches have also gotten increasing attention.
PointNetVLAD~\cite{PR:pointnetvlad} combines the point-based feature extraction and VLAD layer for 3D place recognition. 
LPDNet~\cite{pr:lpdnet} extend~\cite{PR:pointnetvlad} by including local geometric features. 
PCAN~\cite{pcan}, from another perspective, uses the attention-enhanced VLAD layer to improve feature association for accurate localization.
OverlapNet~\cite{PR:overlapnet} provides a differentiable projection layer to estimate the similarity of local 3D sub-maps.
In our previous works, FusionVLAD~\cite{yin2021fusionvlad} provides a fusion based approach to improve the features adaptation between different perspectives; and SphereVLAD~\cite{yin2021fast} provides a viewpoint-invariant place descriptor by combining spherical harmonics~\cite{yin2020seqspherevlad} and sequential matching~\cite{milford2012seqslam}.

Despite the success of existing place recognition methods, the non-learning-based approaches are sensitive to parameter tuning under different scenarios; and learning-based techniques are trained in a supervised learning manner, restricting their generalization ability within the offline training datasets.
However, in real-world localization tasks, the data stream is infinite with the combination of different areas under varying environmental conditions; meanwhile, robotic systems can't stop and wait for the network model to update for newly encountered scenarios.
In this work, we target lifelong learning, where the place observations can be viewed only once in the sequential order~\cite{lifelong_survey}.
In lifelong localization, one important evaluation metric is to analyze the re-localization ability after long-term navigation (i.e., catastrophic forgetting), while most exciting place recognition methods mainly focus on short-term localization or fixed pattern localization~\cite{VPR_Bench}.
To the best of our knowledge, this is the first work to handle real-world long-term/large-scale lifelong localization.

\subsection{Lifelong Learning for Robotics}
\label{sec:lifelong}
Lifelong learning, also known as continual learning, aims at providing incrementally updated knowledge in ever-changing environments.
Though this area has been studied for a long time, most approaches are still restricted to simulation or toy datasets~\cite{lifelong_survey}, and can not be applied in real robotic applications.
As mentioned in~\cite{CLRobot}, the fundamental challenge for lifelong learning is not necessarily finding solutions that work in the real world but rather finding stable algorithms that can learn in the real world and overcome the catastrophic forgetting problem.
Recent works can be roughly divided into four families: dynamic architectures, regularization-based, rehearsal, and generative replay approach.

\begin{figure*}[t]
	\centering
        \includegraphics[width=0.9\linewidth]{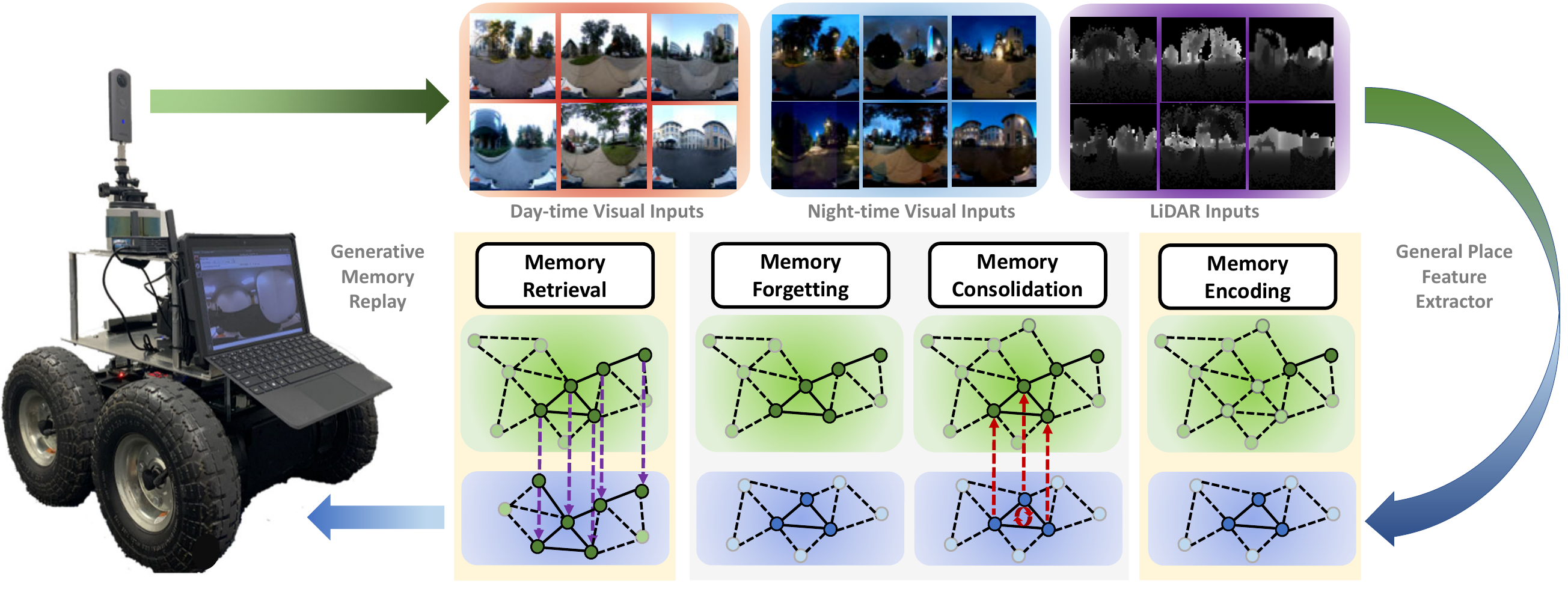}
	\caption{\textbf{Lifelong Localization System Framework.}
	The lifelong localization system contains the following modules:
	1) the general place encoder within the BioSLAM network, which extracts the place feature from different domains;
        2) the dual-memory lifelong learning mechanism within the BioSLAM network can provide short-term and long-term assistance to capture new knowledge  and maintain old knowledge.
    }
	\label{fig:pipeline}
\end{figure*}

Dynamic architecture-based methods either
1)  add additional parameters to the models, such as LwF~\cite{lwf}, which use shared early feature extraction layers and fixed task layers;
or 2) use model adaptation to avoid catastrophic forgetting, such as PackNet~\cite{packnet}, which defines the mask layer to protect weights when learning new tasks.
Regularization-based methods in the context of lifelong learning can add constraints to avoid overfitting to new tasks and keep inference ability for the previous mission, such as Elastic Weight Consolidation (EWC)~\cite{EWC} and Synaptic Intelligence (SI)~\cite{SI}.
However, the above methods must deal with specific network structures and can quickly converge to undesired local optima for complex tasks.
Rehearsal-based methods, on the other hand, use memory replays to enhance the knowledge from the previous tasks or processes such as iCaRL~\cite{icarl}, GEM~\cite{gem}, which use a small subset of the previous dataset to balance the knowledge distribution for different tasks.
Instead of maintaining the knowledge based on past data samples, generative replay~\cite{GP} combines the actual raw data and generated artificial data for model updating.
In~\cite{choi2021dual}, the authors use a dual teacher-student generative replay method for incremental learning, where the teacher network is frozen to guide new networks, and the networks will switch the role when the student network surpasses the teacher.

Hence, the ideal approach would be tackling the real-world localization problem in an embodied platform: an autonomous agent that can efficiently and incrementally update its localization ability with limited computation resources.
Our BioSLAM method combines rehearsal, general replay mechanisms, and a specific dynamic and static memory to tackle long-term complex environments.

    \section{Problem Formulation \& System Overview}
\label{sec:system_oveview}

In this work, BioSLAM represents an incremental place recognition method, which includes:
1) a general place feature extraction module to encode place features under different domains, and 2) the bio-inspired lifelong memory system for online adaptation in place recognition.
In this section, we will first formulate the problem in lifelong localization, then briefly introduce the two key modules in the BioSLAM system.

\subsection{Problem Formulation}
\label{sec:formulation}
We define a sequence of place observations under domain $D$ (i.e., visual, LiDAR, etc.) as $O^{D}=\{O_1,..., O_M\}$, and a query of observations under the same domain as $Q^{D}=\{Q_1,..., Q_N\}$.
The task of traditional place recognition is to learn a feature extraction function $\mathcal{F}$ with parameter $\theta$ to help each frame in $Q^{D}$ find the matched (positive) place from the reference sets $O^{D}$. Let $d(\cdot,\cdot)$ denotes the difference matrix (i.e. Euclidean distance). 
The objective is  to make the feature differences of positive places (or matched) smaller than negative places (or unmatched) by feature extraction function $\mathcal{F}_{\theta^\ast}$.
\begin{align}
\mathcal{L} (Q^D_k) &=  d\left( \mathcal{F}_{\theta}(Q^D_k), \mathcal{F}_{\theta}(O^D_{\approx}) \right)  - d\left( \mathcal{F}_{\theta}(Q^D_k), \mathcal{F}_{\theta}( O^D_{\neq}) \right) \nonumber \\
\theta^\ast &= \arg \min_\theta \sum_{k=1}^N \mathcal{L} (Q^D_k) 
\label{eq:basic_pr}
\end{align}
where $Q^D_k$ is the current $k$-th query, $O^D_{\approx}$ is the positive reference within a predefined neighbor range (e.g. 3m) near $Q^D_k$, and $O^D_{\neq}$ is the negative reference away from $Q^D_k$ more than a predefined threshold (e.g. 10m). 

In lifelong localization, both reference set $O^D_t$ and query set $Q^D_t$ are obtained incrementally, and the environmental domains can also be varying under different environmental conditions (illuminations, weathers, etc.) or sensor modalities.
As depicted in \cref{fig:pipeline}, the lifelong localization problem is to incrementally learn and update the feature extraction function $\mathcal{F}_{\theta}$, that can quickly adapt its feature extraction ability in the newest domain $\{O^{D_T}, Q^{D_T}\}$, and also in parallel maintain the feature distinguish ability for previous domains $\{O^{D_t}, Q^{D_t}\} |_{t=1,...,{T-1}}$. Since we are considering continual learning~\cite{lifelong_survey}, raw data of different domains are fed sequentially for one-time usage and cannot be stored for offline training. Thus, when optimizing feature extraction function $\mathcal{F}_{\theta_T}$ at the current domain  $\{O^{D_T}, Q^{D_T}\}$, we cannot access the previous raw data from $\{O^{D_t}, Q^{D_t}\} |_{t=1,...,{T-1}}$. 
Recall \cref{eq:basic_pr}, for time step $T$,  lifelong localization can be formulated as,
\begin{align}
\mathcal{L}^t (Q^{D_t}_k) &=   d\left( \mathcal{F}_{\theta}(Q^{D_t}_k), \mathcal{F}_{\theta}(O^{D_t}_{\approx}) \right)  - d\left( \mathcal{F}_{\theta}(Q^{D_t}_k), \mathcal{F}_{\theta}( O^{D_t}_{\neq}) \right) \nonumber \\
\theta_T &= \arg \min_\theta  \sum_{t=1}^T  \sum_{k=1}^N \mathcal{L}^t (Q^{D_t}_k)
\label{eq:problem_define}
\end{align}

\subsection{General Place Feature Extraction}
\label{sec:general_place_descriptor}
For the lifelong purpose of long-term localization, we developed a General Place Descriptor (GPD) based on our previous works in visual~\cite{i3dloc} and LiDAR-based~\cite{yin2020seqspherevlad} localization, which can be referred to as the feature extraction function $\mathcal{F}_\theta$ as in \cref{eq:problem_define}.
We use the shared spherical convolution network to achieve LiDAR and visual place localization simultaneously.
The spherical harmonic-based convolution can help the learned descriptor have the viewpoint-invariant propriety for the same place recognition.
The major difference between the current GPD and our previous works is that GPD does not contain any domain-transfer module, which has been used to reduce the feature differences for the same areas under different domains~\cite{i3dloc}.
This modification is because we want to evaluate the adaptation ability for the same network.
On the other hand, there are no task-specific network layers as used in the dynamic architecture-based lifelong modules as stated in \cref{sec:lifelong}.
We want to avoid uncertain parameters and only focus on how memory mechanisms can help incremental learning for real-world applications.

\begin{figure*}[t]
	\centering
    \includegraphics[width=\linewidth]{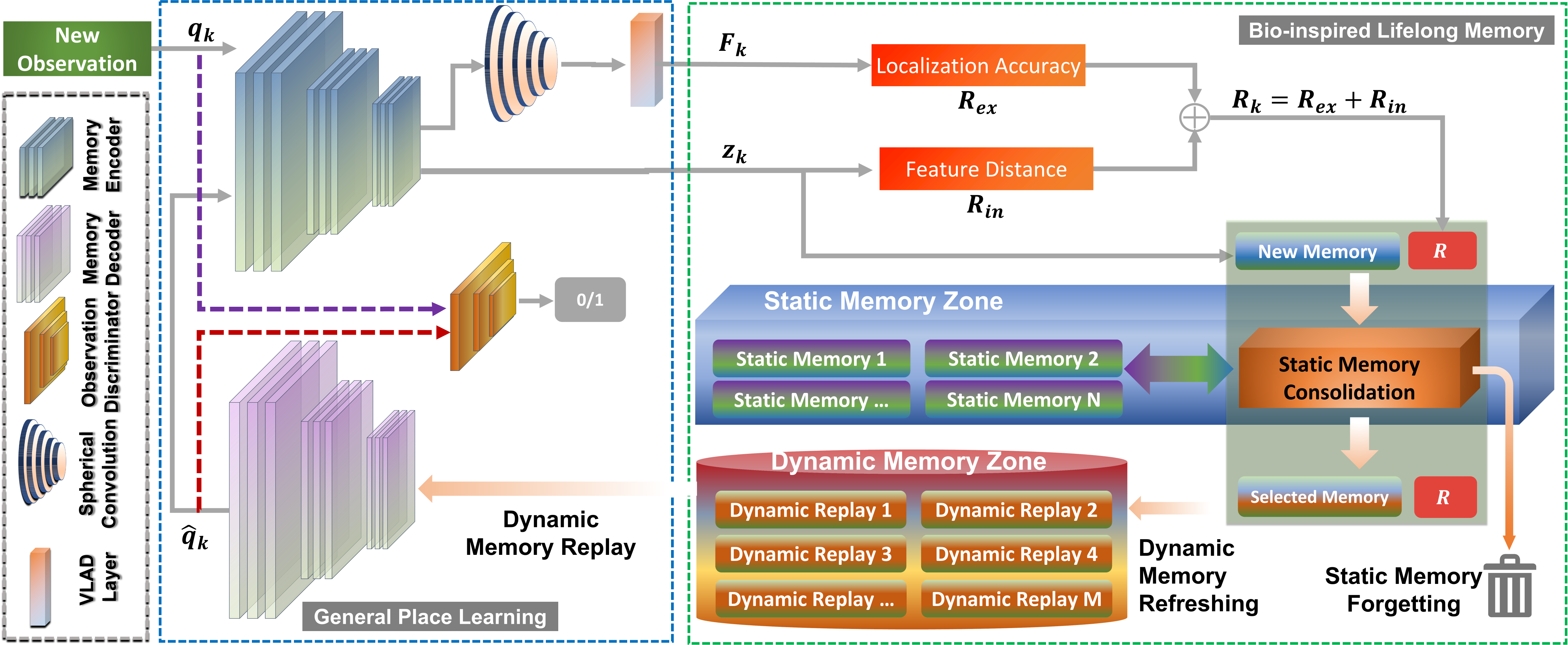}
	\caption{\textbf{BioSLAM Network Structure.}
    The structure of BioSLAM includes the General Place Learner (GPL) network, the rewarding mechanism to guide the memory storing and consolidation, and the dual-memory module with static-/dynamic- memory zones.
    The procedure of new memory encoding includes the following procedure:
    1) new observations $q_k$ are fed into the networks for only one time and in a sequential manner;
    2) In GPL, the memory encoder converts the inputs  $q_k$ to encoded memory $z_k$, followed by spherical convolution and VLAD layer to generate place feature descriptor $F_k$.  
    3) the rewarding mechanism will estimate the external reward $\mathcal{R}_{ex}$ and internal rewards  $\mathcal{R}_{in}$ to guide the memory operations;
    4) the dual-memory will conduct memory storing/consolidation and retrieve (replay) the important (high-rewarded) memory for generative replay.
    5) updating the GPD network module with both observations and replayed memories. 
    }
	\label{fig:network}
\end{figure*}

\subsection{Bio-inspired Lifelong Memory}
\label{sec:system_memory} 
Inspired by the memory system in human-being and other mammal animals~\cite{hippo_memory}, we provide a dual-memory (i.e., dynamic memory zone (DMZ) and static memory zone (SMZ)) enhanced lifelong learning mechanism to deal with catastrophic forgetting in continual localization.
As studied in~\cite{stickgold2005sleep}, to create long-term memories in our brain, we have so-called \textit{sleep circle} during our sleep: 
\begin{itemize}
    \item 1) the brain can \textit{encoding} our daily observation into the \textit{hippocampus} zone with decay along the time; 
    \item 2) then, a consolidation mechanism is triggered between the hippocampus and the neocortex to store essential memory traces and forget the rest traces;
    \item 3) finally, humans can retrieve the relative memory traces based on the consolidated ones in the neocortex.
\end{itemize}
In BioSLAM, we re-build the `\textit{sleep circle}' for the lifelong localization task.
As we can see in \cref{fig:pipeline}, the memory system of BioSLAM also includes the place feature `encoding' procedure for new observations, the memory `consolidation' controlled by a behavior cost module to filter out necessary traces for more extended storage, and the `retrieved' memory to re-enhance the long-term place recognition ability.
Based on the above architectural, BioSLAM system construct two major systems, the \textit{General Place Learning} (GPL) system and the \textit{Bio-inspired Lifelong Memory} (BiLM) system, which will be investigated in \cref{sec:GPFL} and \cref{sec:BiLM} respectively.
    \section{General Place Learning}
\label{sec:GPFL}
As shown in \cref{fig:network}, the general place learning (GPL) (blue dashed box) system mainly contains two sub-modules: a place memory encoding module (upper part of the blue dashed box) and a generative memory reply module  (lower part).
All the data under different domains (LiDAR inputs or visual inputs on day/night conditions) are fed into the system sequentially once during the online training procedure.
The GPL system uses the symmetric encoder-decoder networks to encode new observations and decode memory traces.
In this section, we introduce the design of the encoder, the decoder, and the place feature learning within the GPL system and leave the memory system in the next section.

\subsection{Place Memory Encoding}
\label{sec:encoding}
GPL applies the \textit{encoder module} $\mathcal{E}$ to convert raw sensor observations into the `memory codes' with VGG~\cite{Feature:VGG}-based networks, which are also the basic materials in the BiLM system.
In parallel, GPL constructs the \textit{decode module} mirrored to the encode module, which can reconstruct the stored memory into the \textit{synthetic observations}.
Since in lifelong localization, both viewpoint difference and environmental appearance changes will affect the final localization performance in real-world applications. 
Based on orientation-equivalent property of spherical harmonics, we utilize the spherical convolution~\cite{i3dloc,yin2020seqspherevlad} in the \textit{encoder module} to provide viewpoint-invariant descriptor to reduce the viewpoint differences in long-term re-localization.
As shown in \cref{fig:network}, the extracted place feature descriptor is not involved with the BiLM memory system, which indicates that our lifelong localization is mainly designed for the long-term domain differences, which also helps simplify quantitative and qualitative analysis.

The GPL system  encodes both panorama camera and 3D local point cloud with the same encoding network structure $\mathcal{E}$.
For the visual inputs, we convert the raw image to $[H\times W]$ spherical perspectives, which are fed into the encoder module for memory storage and place descriptor extraction.
For the LiDAR inputs, instead of a single scan, we generate dense local 3D maps using the similar voxel mapping mechanism in our previous work~\cite{yin2021fusionvlad} and map the points onto the spherical projections, which have the same omnidirectional view as a panorama camera.
The default size for visual and LiDAR views is $H=W=64$, with $3$ channels for visual and LiDAR inputs (repeated three depth channels ranging from $0$ to $1$).
We can obtain the encoded `memory' $z_k$ from observation $q_k$ by,
\begin{align}
    z_k = \mathcal{E} (q_k) \label{eq:encoder}
\end{align}

To extract the orientation-invariant place descriptor from $z_k$, we utilize the spherical convolution based on the spherical harmonics~\cite{Sphere:SO3_learning}.
In theory, Spherical convolution can avoid space-varying distortions in Euclidean space by convolving spherical signals in the harmonic domain. 
Let $f$ is the signal on spherical harmonic, which satisfy the orientation-equivariant~\cite{cohen2018spherical} property with the signal $\mathcal{E}$, 
\begin{align}
    \left[f \star_{SO(3)} [H_{R} \mathcal{E}] \right] (q_k)
    = & \left[ H_R[f \star_{SO(3)} \mathcal{E}] \right](q_k)
\end{align}
where $H_R \: (R \in SO(3))$ is the rotation operator for spherical signals. $f \star_{SO(3)} \mathcal{E}$ denotes the spherical convolution between $f$ and $\mathcal{E}$. Practically, the spherical convolution is computed in three steps. We first expand $f$ and $\mathcal{E}(q_k)$ to their spherical harmonic basis, then compute the point-wise product of harmonic coefficients, and finally invert the spherical harmonics.

Let $V$ be the unsupervised VLAD layer~\cite{NetVLAD}. Then the
feature extraction function  $\mathcal{F}_\theta  =V  \circ [ f \star_{SO(3)} \mathcal{E}]$  is the orientation-invariant function. Where $\theta$ are learnable parameters of the feature extraction function. 
Given the data sample $q_k$, the place descriptor (or learned feature) $F_k$ can be denoted as
\begin{align}
   F_k= \mathcal{F}_\theta (q_k)  = 
   V  \circ [f  \star_{SO(3)} z_k ]
    \label{eq:lean_feature}
\end{align}
For details about the viewpoint-invariant analysis, please refer to our previous works~\cite{i3dloc,yin2020seqspherevlad}.

The above procedure is relevant to the biological `encoding' procedure within the `sleep cycle' as we mentioned \cref{sec:system_oveview}, and the extracted `memory codes' $z_k$ will be used for later `memory consolidation' in next \cref{sec:BiLM} and `retrieval' for generative replay in next \cref{sec:generative_replay}

\subsection{Generative Memory Replay for Place Recognition}
\label{sec:generative_replay}
As depicted in \cref{fig:network}, in our generative memory replay for the lifelong localization task, the data stream under different environmental conditions is fed into the system sequentially, and we generate synthetic samples from stored place memories through a deep generative memory replay framework.
In particular, the retrieved `memories' will include the abstracted place latent codes under different domains as shown in \cref{fig:replay}, which enforce the generative memory play to extract a portion of history samples under all different domains to maintain the localization performance.
The next section relates the memory extraction mechanism to the BiLM memory management.

To ensure the generalization ability of synthetic samples, we provide a deep generative adversarial network (GANs) to mimic the distribution differences between the raw data and synthetic samples and parallel with a $L1$ reconstruction loss between encoder and decoder modules. 
The GANs-based generative model defines a zero-sum minimax game with the memory decoder $\mathcal{G}$ and the discriminator $\mathcal{D}$ as stated in~\cite{cnn:gan}, the objective function is thereby defined by,
\begin{align}
 & \min_{\mathcal{G}} \max_{\mathcal{D}} \mathcal{L}_{gan} (\mathcal{G}, \mathcal{D}) = \\
   & \min_{\mathcal{G}} \max_{\mathcal{D}}  \mathbf{E}_{q\sim P_{data}}[\log \mathcal{D}(q)] + 
    \mathbf{E}_{z' \sim P_{z}}[\log (1-\mathcal{D}(\mathcal{G}(z')))]   \nonumber
    \label{eq:replay_sample}
\end{align}
where $P_z$ is the retrieved memory buffer from the BiLM system, and $P_{data}$ is the new observed data samples. $\mathcal{L}_{gan}$ denotes the generator and discriminator losses \cite{cnn:gan}.
The detailed generative play strategy for lifelong learning can be found in~\cite{Shin2017deepreplay,vandeven2020brain}.

Given the combination of both new streaming data and retrieved synthetic samples $\{P_{data}, P_z\}$, we can obtain observation pair $\{Q, O\}$
with tuple sets $(q_k, \{ o^{pos}_k \}, \{o^{neg}_k\})$, where for each query sample $q_k$ we have a set of potential positives (close-by samples ) $\{ o^{pos}_{k} \}$ and the set of negatives (far away samples) $\{  o^{neg}_{k}  \}$. 
The localization loss metric is defined by:
\begin{align}
&L_{loc}(q_k) = \\
\max_{i,j} &\left( \|\mathcal{F}(q_k) -\mathcal{F}({o^{pos}_k}_i)\|^2 + \alpha - \|\mathcal{F}(q_k) -\mathcal{F}({o^{neg}_k}_i)\|^2 , 0 \right) \nonumber \\
&\mathcal{L}_{loc} = \mathbf{E}_{q_k \sim P_{data}} [L_{loc}(q_k)] + \mathbf{E}_{z' \sim P_z} [L_{loc}(\mathcal{G}(z'))]
\label{eq:loc_loss}
\end{align}
the above equation is the triplet loss version of \cref{eq:basic_pr}, where $L_{loc}(q_k)$ is the localization loss metric for single query $q_k$, ($\alpha > 0$) is a margin to control the feature difference threshold, and $\mathcal{L}_{loc}$ is the localization loss for the joint $\{P_{data}, P_z\}$ sets.

\begin{figure}[t]
	\centering
    \includegraphics[width=\linewidth]{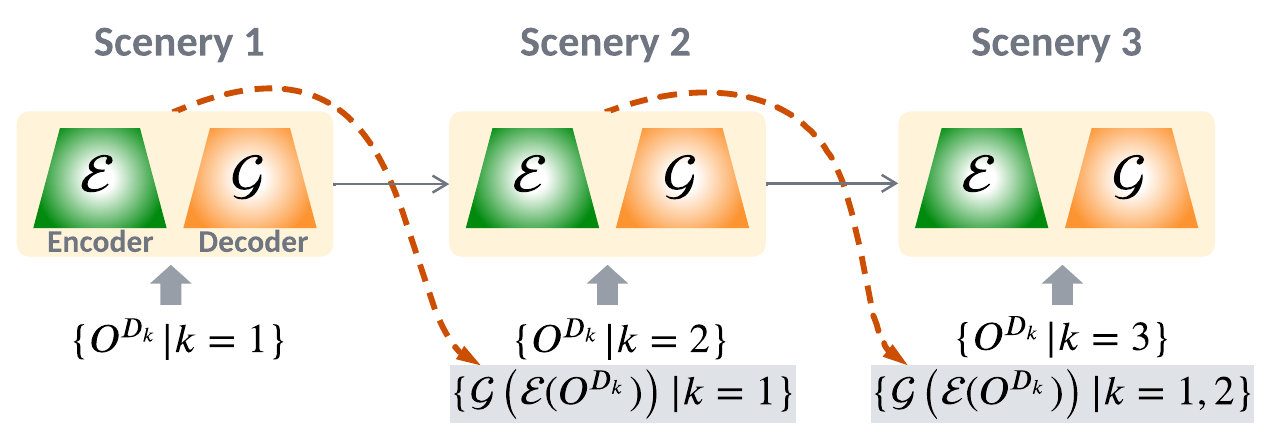}
	\caption{\textbf{Generative Memory Reply within the GPL system.}
	In the lifelong localization, new observations $O^{D_t}$  under domains $D_t$ will be streamed into the BioSLAM system sequentially.
	GPL's generative memory replay module can generate synthetic samples $\mathcal{G}\left(\mathcal{E}(O^{D_t})\right)$ from stored memories.
    }
	\label{fig:replay}
\end{figure}

To keep the consistency of memory encoding-decoding, we further use a reconstruction loss of the memory $z'$ and the generated memory $\mathcal{E}(\mathcal{G}(z'))$ with,
\begin{align}
  \mathcal{L}_{rec} =  \mathbf{E}_{z' \sim P_z} [\|\mathcal{E}(\mathcal{G}(z')) - z' \|]
\end{align}
And the joint loss metric for the generative memory replay enhanced place recognition can be written as,
\begin{align}
  \mathcal{L}_{joint} = \mathcal{L}_{loc} + \mathcal{L}_{rec} + \mathcal{L}_{gan} (\mathcal{G}, \mathcal{D})
  \label{eq:loss_joint}
\end{align}

The major difference between our work and the traditional generative play~\cite{Shin2017deepreplay} is that BioSLAM can manage the retrieved memory based on their long-term behavior instead of treating all data on the same manifold distribution. 
The next section will deeply investigate the lifelong memory system.
    \section{Bio-inspired Lifelong Memory}
\label{sec:BiLM}

As we analyzed in \cref{sec:lifelong}, most current lifelong learning methods target toy examples, which can not be generalizable under complex real-world environments.
In our BioSLAM,  as shown in the greed dashed box of \cref{fig:network}, the bio-inspired lifelong memory system mainly contains two modules: 
1) a \textit{behavior configurator} module to arrange memory consolidation and selection, which is based on memory traces' importance (measured by reward calculation) to the long-term place recognition.
and 2) a \textit{dual-memory} module to cooperate with the behavior configurator for long-/short- term memory storage and importance-retrieval with limited space usage; which includes static memory zone and dynamic memory zone.

\subsection{Behavior Configurator} 
\label{sec:behave_configure}
When the memory system encounters a new place `memory traces' $z$, we define the hybrid cost to control the learning behavior: an external reward $\mathcal{R}_{ex}$ which indicates localization ability, and the internal reward $\mathcal{R}_{in}$ which can present the intrinsic familiarity on observations.


\subsubsection{External Reward}
\label{sec:ex_reward}
The external reward is related to the learning difficulty of new data samples, which indicates the distinguishing ability in the place recognition task.
In the standard training paradigm, all samples under different difficulty levels are almost equally used to optimize the training model.
However, Humans and animals always spend more energy and time learning more complex concepts. 
Inspired by the animal training \cite{krueger2009flexible} and curriculum learning \cite{bengio2009curriculum}, it is practically useful to sort the data samples into different difficulty level, i.e., ``easy", ``medium" and ``hard".
For lifelong localization, the ``hard" samples may need more `energy' to encode in the training model $\mathcal{F}$, i.e., more retrieval times as in the replay procedure as stated in \cref{sec:generative_replay}.
To encourage the ``harder'' samples to have a higher chance of re-training, we define the triplet loss to measure features' distinguishability.
Based on place recognition loss metric $L_{loc}$, we define the external reward for every single query as,
\begin{align}
    \mathcal{R}_{ex} (q_k) = L_{loc} (q_k)
\end{align}
which means that if query $q_k$ has a higher loss than other queries, it will require more `energy', i.e., more iteration times in model training.
``Harder" samples will tend to higher $\mathcal{R}_{ex}$, then the BiLM memory system will have a higher chance to retrieve them for memory replay as stated in \cref{seq:dynamic_replay}.

\subsubsection{Internal Reward}
\label{sec:in_reward}
The internal reward is related to the robustness of feature representations. Let $A(q_k)$ denote the data augmentation (i.e. random rotation and random translation) for query  $q_k$.  
The internal reward $\mathcal{R}_{in}$ for query $q_k$ is defined by the cosine distance of features between its augmented version,
\begin{align}
& \mathcal{R}_{in}(q_k) = 1 - \frac{\mathcal{E}(q_k) \cdot \mathcal{E}(A(q_k))}{ \|\mathcal{E}(q_k)\|_2 \cdot \|\mathcal{E}(A(q_k))\|_2 }
\label{eq:reward_in}
\end{align}

The internal reward  $\mathcal{R}_{in}$ also indicates the network's familiarity with the observations. 
This is common in large-scale structured areas, such as LiDAR-based place recognition under city-scale environments.
In that case, similar place patterns (street view, buildings, trees) can be frequently visited with different views; the $\mathcal{E}$ has a robust representation and lower internal reward of frequently visited places.  Thus, the inner reward $\mathcal{R}_{in}$ can be applied as an indicator to guide the memory system on whether or not to pay more attention to such areas. 
The inner reward can provide intrinsic property analysis for the memory encoder based on the above analysis.

The final reward for $q_k$ can be obtained by combining the external reward with the internal reward,
\begin{align}
 \mathcal{R}_k =  \mathcal{R}_{ex}(q_k)+\mathcal{R}_{in}(q_k),
 \label{eq:reward_total}
\end{align}
Based on this rewarding mechanism, we can evaluate all the queries, and obtain a set of memory trace $m_k = (z_k, p_k, \mathcal{R}_k)$, where $p_k$ is estimated location of $q_k$ through our previous re-localization system~\cite{i3dloc,yin2022automerge}.
$m_k$ is then the main factor used in memory operations of \cref{sec:memory}.

\subsection{Dual-Memory \& Memory Operations}
\label{sec:memory}
The memory of human beings is highly connected with long-term memory (the neocortex) and short-term memory (the hippocampus) mechanisms within our brains.
BioSLAM also constructs such paired dual-memory mechanisms, 
\begin{itemize}
    \item \textbf{Static Memory $M_S$} is similar to the long-term memory of human beings and belongs to rehearsal-based mechanisms with large storage for lifelong learning. Static memory stores the selected memory traces $\{m_k\}$ to the static memory zone by memory consolidation.
    \item \textbf{Dynamic Memory $M_D$} is similar to the short-term memory of human beings, which  is a quick access memory with a portion of pre-stored historical memory traces. Dynamic memory is automatically refreshed from the static memory and connected with the memory decoder module, which belongs to generative replay mechanisms with small memory buffer for lifelong learning.
\end{itemize}
Based on the dual-memory structure, we construct two important operations for static memory: memory consolidation and forgetting, and  two important operations for dynamic memory: memory refreshing and memory replay. 

\begin{algorithm}[t]
    \KwInput{Static memory $M_S$, new memory traces $\{m_k\}$, maximum number of clusters $K_{max}$ }
    \KwOutput{Updated static memory}
    Construct feature-spatial codes $\{c_k\}=\{z_k,p_k|m_k\}$\;
    Calculate clusters $\{S_i^T\}|_{i=1}^K$ and centroids $\{\mu_i^T \}|_{i=1}^K$ for $\{c_k\}$ based on \cref{eq:clusters}\;
    Downsample within clusters, based on \cref{eq:downsample}, to generate smaller clusters $\{\tilde{S}_i^T\}|_{i=1}^K$ and centroids $\{\tilde{\mu}_i^T \}|_{i=1}^K$ \;
    Append new clusters $\{\tilde{S}_i^T\}|_{i=1}^K$ and centroids $\{\tilde{\mu}_i^T \}|_{i=1}^K$ to  $M_S$\;
    Calculate the total cluster number $C^{(M_s)}$ in  $M_S$ \;
    \If{ $ C^{(M_s)} > K_{max}$} 
    {
        Memory Forgetting with \cref{algo:mem_forget}\;
    }
    Update static memory $M_S$\;
    \caption{Memory Consolidation}
    \label{algo:consolidation}
\end{algorithm}

\subsubsection{Static Memory Consolidation}
\label{sec:mem_consolidation}

As stated in~\cite{byrne2017learning}, memory consolidation is defined as a time-dependent process by which recently learned experiences are transformed into long-lasting forms to extend the long-term memory circle.
In the long-term and large-scale place recognition task, the observations may include differences in the spatial domain (Euclidean distance) and feature domain (feature distance). The data stream is also unlimited in the real-world navigation task.
Memory consolidation is essential to abstract concise representations and guarantee memory efficiency.
To provide memory consolidation within BioSLAM system, we construct a feature-spatial code $c_k=[z_k,p_k]$ for memory trace $m_k$, which can capture both spatial and feature properties.

Given time step $T$ and observations $\{q_k\}^T \subset Q^{D_T}$, the obtained new memory traces $\{m_k\}^T=\{(z_k, p_k, \mathcal{R}_k)\}^T$ ( \cref{eq:encoder,eq:reward_total},) usually contains a large number of samples. We get the diverse and smaller subset of memory traces (abstraction) via K-means-based unsupervised clustering.  K-means clustering partition the $\{m_k\}^T$ into $K$ sets $\mathbf{S}^T=\{S_1^T, S_2^T, \cdots, S_K^T \}$ by feature-spatial code $c_k=[z_k,p_k]$ to minimize the following,
\begin{align}
    \mathbf{S}^T &= \argmin_{\mathbf{S}^T} \sum_{i=1}^{k} \frac{1}{|S_i^T|} \sum_{c_x,c_y\in S_i^T}\|c_x-c_y\|^2
    \label{eq:clusters} \\
    \mu_i^T &=  \frac{1}{|S_i^T|} \sum_{c_i\in S_i^T} c_i \nonumber
\end{align}
where $\mu_i^T$ is the cluster centroid for cluster $S_i^T$. The memory traces of a cluster can be served as mutually homogeneous. Storing all memory traces from a cluster is redundant and memory exhaustive. 
Thus down-sampling is used within clusters $S_i^T$ to restrict the number of samples for each cluster, which is also helpful in improving the memory retrieval efficiency. 
\begin{align}
    (\tilde{S}_i^T, \tilde{\mu}_i^T) = { \rm downsampling}(S_i^T, \mu_i^T), \quad  |\tilde{S}_i^T|<N_{max}
    \label{eq:downsample}
\end{align}
Where $N_{max}$ is the predefined threshold for the maximum number of samples in each cluster. 
After sampling, we generate $K$ smaller clusters $\mathbf{\tilde{S}}^T=\{\tilde{S}_1^T, \tilde{S}_2^T, \cdots, \tilde{S}_K^T \}$ and centroids $\mathbf{\tilde{\mu}}^T=\{\tilde{\mu}_1^T, \tilde{\mu}_2^T, \cdots,\tilde{\mu}_K^T \}$ from subset of new traces $\{m_k\}^T$. 

BioSLAM can combine the current new clusters $\mathbf{\tilde{S}}^T$ with the existing clusters from previous steps, then the total clusters in the static memory are  $\mathbf{S}^{(M_s)} = \{\mathbf{\tilde{S}}^1, \mathbf{\tilde{S}}^2,\cdots, \mathbf{\tilde{S}}^T\}$, and the centroids are $\mathbf{\mu}^{(M_s)} = \{\mathbf{\tilde{\mu}}^1, \mathbf{\tilde{\mu}}^2,\cdots, \mathbf{\tilde{\mu}}^T\}$. 
When the total cluster number $C^{(M_s)} = |\mathbf{\mu}^{(M_s)}|$ is beyond the maximum threshold $K_{max}$, 
some similar clusters are merged to avoid the memory overflow described in the Memory Forgetting \cref{sec:mem_forget}. 
The consolidation mechanism is shown in \cref{algo:consolidation}.

\begin{algorithm}[t]
  \KwInput{Static memory $M_S$, maximum number of clusters $K_{max}$}
  Load clusters $\mathbf{S}^{(M_s)}$ and centroids $\mathbf{\mu}^{(M_s)}$ from static memory $M_S$ \;
  Calculate the number of forgettable clusters $K^\ast =|\mathbf{\mu}^{(M_s)}| -K_{max} $ \;
  Calculate the distance matrix $d_{(i,j)}$ between every two clusters based on \cref{eq:memory_dist} \;
  \While{ {\rm repeat} $K^\ast$ {\rm times}}
  {
    Find most similar cluster pairs $(i^\ast, j^\ast)$ based on \cref{eq:memory_sim} \;
    Remove cluster $i^\ast$ from static memory $M_S$ and distance matrix $d_{(i,j)}$ \;
  }
 \caption{Memory Forgetting}
 \label{algo:mem_forget}
\end{algorithm}

\subsubsection{Static Memory Forgetting}
\label{sec:mem_forget}
As stated in the last section, the space within static memory is bounded in long-term lifelong learning.
As a core operation in static memory, memory forgetting is designed to eliminate the redundant memory clusters when they are too similar to the other existing clusters.
If the current cluster number is bigger than the maximum number of clusters  $C^{(M_s)} > K_{max}$, memory forgetting mechanisms remove number of $K^\ast =C^{(M_s)}-K_{max} $ clusters. 
We first calculate the cluster similarity based on the distance matrix $d_{(i,j)}$ between every two cluster centroids. 
\begin{align}
    d_{(i,j)} =\| \mu_i - \mu_j \|, ~ \forall \mu_i,\mu_j \in \mu^{(M_s)}
    \label{eq:memory_dist}
\end{align}
Then we find corresponding cluster pairs $(i^\ast, j^\ast)$ with the minimum distance and remove one of the clusters $i^\ast$ from the selected pairs.  
\begin{align}
    (i^\ast, j^\ast) =\argmin_{i,j} d_{(i,j)}
    \label{eq:memory_sim}
\end{align}
The removal process will be repeated $K^\ast$ times.
In this manner, we can efficiently keep the diversity of memory clusters and eliminate the `redundant' clusters. The memory forgetting mechanism is shown in \cref{algo:mem_forget}.


\subsubsection{Dynamic Memory Refreshing}

Dynamic memory is brief and storage-limited, just like the short-term memory of humans. In order to effectively replay important memory traces from dynamic memory, we need to refresh dynamic memory and convert memory traces from static memory to dynamic memory at some frequency. In memory refreshing mechanisms, dynamic memory $M_d$ obtain memory traces $\{m_k\}$ from static memory $M_s$ by importance sampling,
\begin{align}
    M_d &= {\rm importance\_sampling}(\{m_k\},\{ w_k \})    \label{eq:mem_refresh}  \\
    m_k &= (z_k, p_k, \mathcal{R}_k) \sim M_s, \quad w_k = \gamma^{n(m_k)} \cdot \mathcal{R}_k  \nonumber
\end{align}
where importance weights $w_k$ are determined by the reward $\mathcal{R}_k$ and the time-decaying factor $\gamma^{n(m_k)}$. $\gamma$ ($0 \le \gamma \le 1$) is a predefined decay parameter, and $n(m_k)$ denotes the replayed time (or revisited time) for the trace $m_k$. On the one hand, traces with higher rewards have higher sampling weights. Because higher rewards mean lower localization ability and robustness, BioSLAM need to pay more attention to these samples. On the other hand, new traces have higher sampling weights. Because the network's ability to learn samples with many occurrences has reached an upper limit, there is no need to 
spend precious dynamic memory to store samples that have been replayed many times. The decaying mechanisms also encourage dynamic memory to increase curiosity about new traces. 
The above reward decay mechanisms are inspired by the decaying factor in human memory \cite{berman2009search}, which indicates that repeated learning of the same things will decrease the boost in memorization.


\subsubsection{Dynamic Memory Replay}
\label{seq:dynamic_replay}
During lifelong learning, BioSLAM retrieves memory traces from dynamic memory for the generative replay training as stated in \cref{sec:generative_replay}. 
In dynamic memory replay, we still use importance sampling to obtain replayed memories $\{z'_k \}$ from dynamic memory $M_d$ with the same as the refreshing memory mechanisms.
\begin{align}
    \{z'_k \} &= {\rm importance\_sampling}(\{z_k\},\{ w_k \})  \label{eq:mem_replay} \\
    m_k& =(z_k, p_k, \mathcal{R}_k) \sim M_d, \quad w_k = \gamma^{n(m_k)} \cdot \mathcal{R}_k \nonumber
\end{align}
Then we use memory decoder $\mathcal{G}$ to generate replayed samples $\{ \hat{q}_k\}$ from memories $\{z'_k\}$,
\begin{align}
   \hat{q}_k =  \mathcal{G}(z'_k)
   \label{eq:decode}
\end{align}
Both new observations $\{q_k \}$ and generated samples  $\{ \hat{q}_k\}$ are used to train General Place Learner (GPL) network with minimizing the total loss \cref{eq:loss_joint}. The overall lifelong learning algorithm of BioSLAM is shown in \cref{algo:bioslam}. 

\begin{algorithm}[t]
  \KwInput{Initial place feature extraction model $\mathcal{F}_\theta$ with parameters $\theta$, Initial static memory $M_s=\emptyset$ and dynamic memory $M_d=\emptyset$}
  \For{ $T=1,2,\cdots$ }
  {
   Obtain observation set $\{q_k \}$ \;
   \While{ {\rm repeat until converge}}
   {
   Generate replayed samples  $\{\hat{q}_k \}$  from dynamic memory based on \cref{eq:mem_replay,eq:decode} \;
   Calculate loss $ \mathcal{L}_{joint}$ using real samples $\{q_k \}$ and replayed samples  $\{\hat{q}_k \}$ based on \cref{eq:loss_joint} \;
   Calculate gradient $\frac{d \mathcal{L}_{joint}}{d \theta}$ then optimize $\mathcal{F}_\theta$ with parameters $\theta$ by gradient descend \;
   }
   Calculate rewards $\{\mathcal{R}_k\}$ of observations $\{q_k \}$ based on \cref{eq:reward_total} \;
   Static memory $M_s$ consolidation based on \cref{algo:consolidation} \;
   Dynamic memory $M_d$ refreshing based on \cref{eq:mem_refresh} 
  }
 \caption{Lifelong Learning with BioSLAM}
 \label{algo:bioslam}
\end{algorithm}

%

    \begin{figure}[ht]
	\centering
    \includegraphics[width=\linewidth]{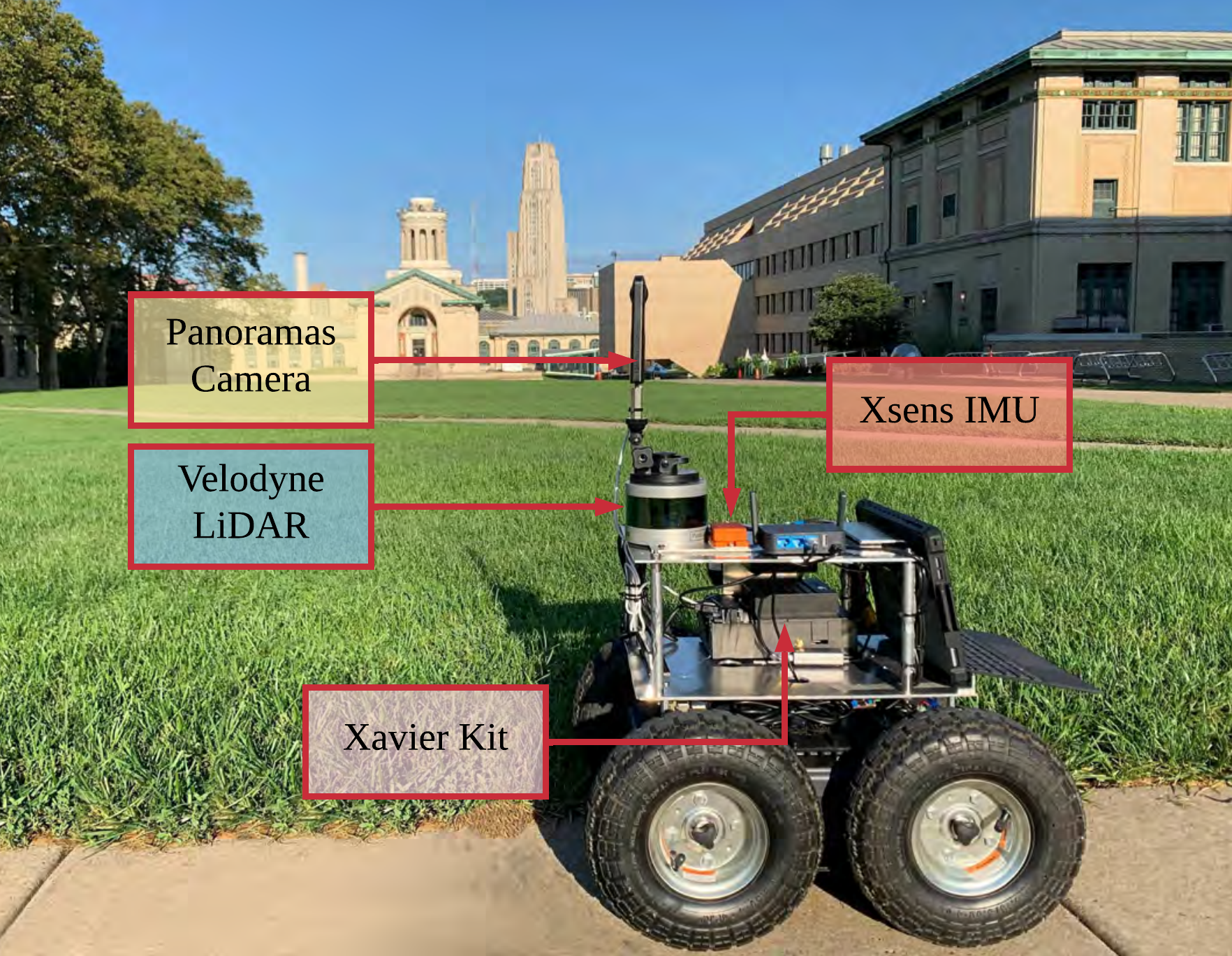}
    \caption{\textbf{Data Collection Platform.}
    The platform can record the omnidirectional visual inputs, Velodyne VLP-16 LiDAR inputs, and Xsens MTI IMU data on an Nvidia Jetson AGX Xavier.
    We utilize the LiDAR odometry~\cite{LOAM:zhang2014loam} to generate the relative odometry for each trajectory and GNSS or Generalized-ICP~\cite{segal2009generalized} to estimate the relative transformation between different trajectories.}
	\label{fig:platform}
\end{figure}

\begin{figure*}[th]
    \centering
    \includegraphics[width=\linewidth]{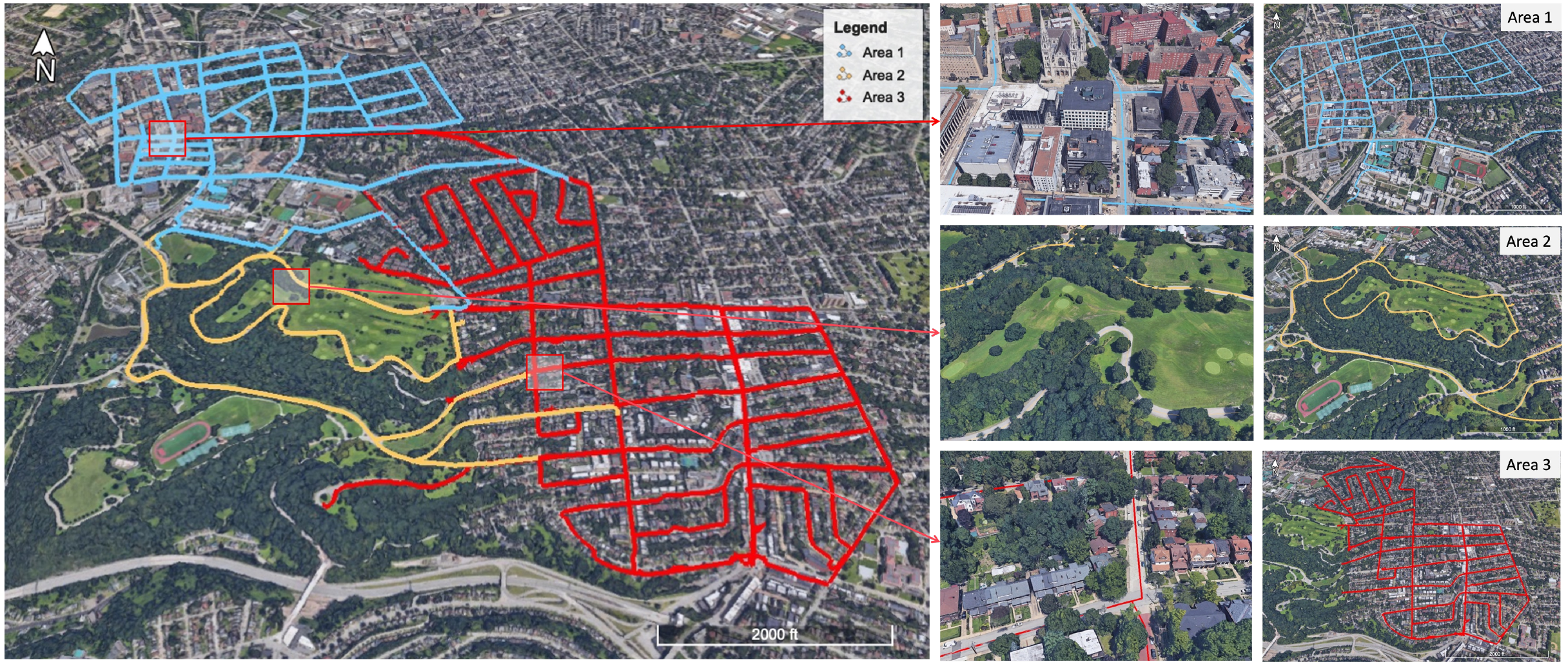}
    \caption{\textbf{City Dataset for Lifelong Localization.}
    The \textit{City} dataset includes 50 trajectories (110 km) within the city of Pittsburgh.
    The dataset includes three areas (colored in blue, yellow, and red) covering commercial buildings, parks, and residential areas.
    }
    \label{fig:pitt_dataset}
\end{figure*}

\begin{figure*}[th]
    \begin{center}
    \includegraphics[width=\linewidth]{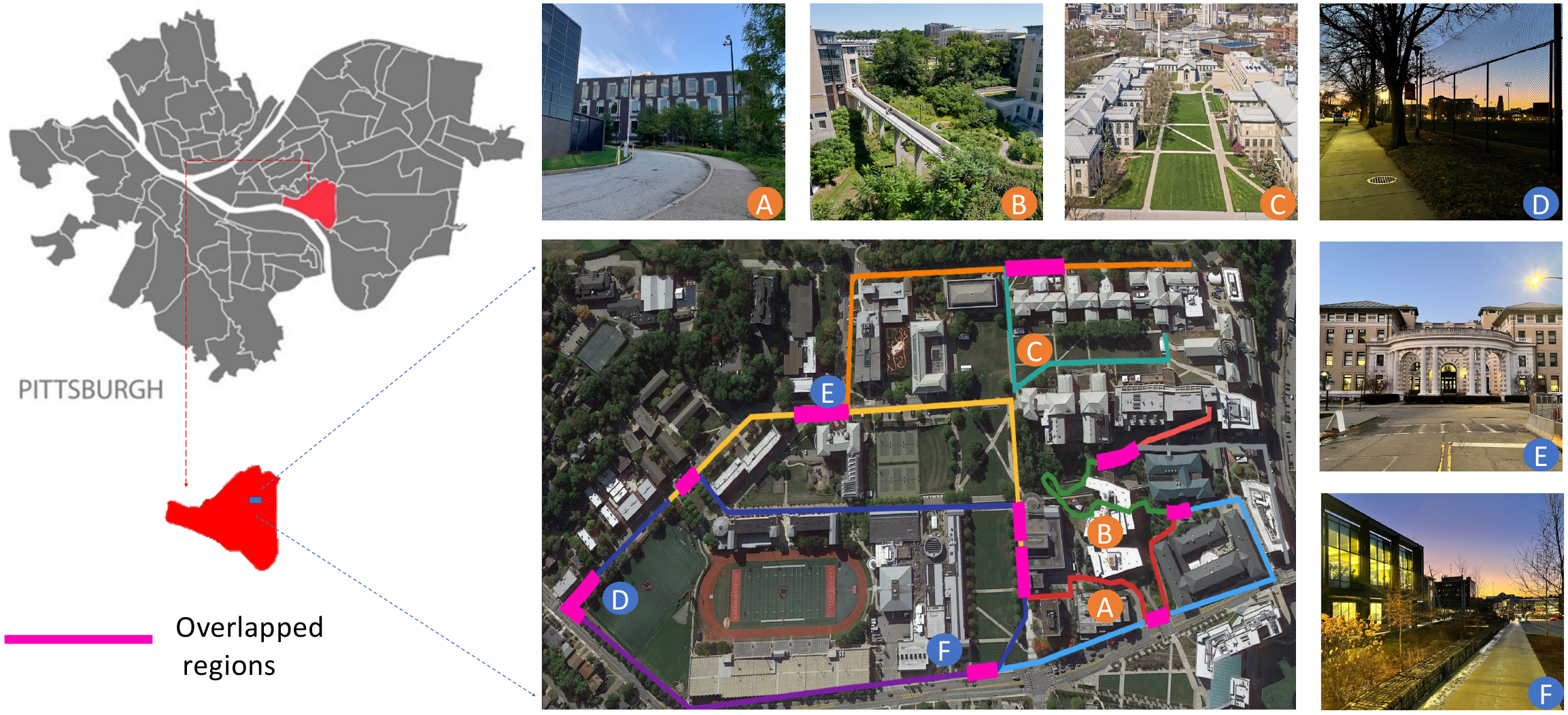}
    \end{center}
    \vspace{-5pt}
    \caption{
    \textbf{Campus Dataset for Lifelong Localization.}
    For \textit{Campus} dataset, omnidirectional camera and LiDAR data are recorded for 2D-to-2D and 2D-to-3D place recognition within CMU.
    The campus datasets are generated during $08/2021 \sim 10/2021$, which are mainly taken from normal day-light ($2pm \sim 5pm$) and dawn-light ($5am \sim 6am$ or $7pm \sim 8pm$).
    }
    \label{fig:campus_dataset}
\end{figure*}

\section{Experiment Setup and Criteria}
\label{sec:exp_setup}
In this section, we will introduce the experiment setup for lifelong localization.
Different from traditional localization tasks, lifelong localization requires the recorded data includes either long-term differences or large-scale geometric differences.
And based on the above reasons, we built our own data collection platform and own lifelong localization datasets. 
We will also briefly describe our evaluation metrics.




\subsection{Data Collection Platform}
\label{sec:data_platform}

\Cref{fig:platform} shows our data collection platform, which includes an omnidirectional camera, a Velodyne VLP-16 LiDAR device, an inertial measurement unit (Xsense MTI $30$, $0.5^{\circ}$ error in roll/pitch, $1^{\circ}$ error in yaw, $550m$W), and an embedded GPU device (Nvidia Xavier, $8$G memory).
To collect time-synced LiDAR projection and omnidirectional images, we first generated dense 3D maps through well-known LiDAR odometry~\cite{LOAM:zhang2014loam}. Then project the point cloud within a certain distance (default is $30m$) to the spherical projections, which have the same perspective as the omnidirectional images.
We will revisit the same area under large-scale and long-term assumptions in lifelong localization. 
To provide the relative ground truth position between different visits: to outdoor environments, we rely on the GNSS system and Generalize-ICP~\cite{segal2009generalized} to estimate the relative transformation;
For indoor environments, we mainly rely on Generalize-ICP.
Please note that we cannot guarantee the meter-level global absolute localization, but we can provide accurate relative localization, which is enough for the lifelong localization task.
Based on the collected datasets, we have hosted a General Place Recognition Competition for long-term place recognition. 
For more details on the data collection platform and the datasets, please refer to our dataset paper(\href{https://github.com/MetaSLAM/ALITA}{https://github.com/MetaSLAM/ALITA}) and competition site (\href{http://gprcompetition.com/}{http://gprcompetition.com/}).

   \begin{table} [t]
        \caption{Comparison between different datasets.}
        \vspace{-5pt}
        \centering
        \begin{tabular}{|C{1.5cm}|C{4cm}|C{1.5cm}|}
            \hline
            \textbf{Dataset}
            & \textbf{Environments}
            & \textbf{Scales (km)}
            \\
            \hline
            City & Street, Residential, Terrain & $120\times 1$\\
            \hline
            Campus & Campus area & $4.5\times 8$\\
            \hline
        \end{tabular}
        \label{tab:dataset}
    \end{table}

\subsection{Lifelong Localization Datasets}
\label{sec:dataset}
We intend to analyse the lifelong performance under \textit{large-scale} and \textit{long-term}  two perspectives.
To this end, our localization datasets include two tracks:
\begin{itemize}
    \item \textit{City} dataset: shown in \cref{fig:pitt_dataset}, which is targeting at large-scale lifelong performance.
    We collected $50$ trajectories within the city of Pittsburgh.
    Since we mainly care about large-scale localization, we only collected the LiDAR inputs within a short-term drive.
    The total trajectory distance for this dataset is $110$km.
    \item \textit{Campus} dataset: shown in \cref{fig:campus_dataset}, which is targeted at long-term lifelong performance.
    We picked up $10$ trajectories within Carnegie Mellon University. 
    Each trajectory is revisited by $8$ times under different day- and night- time to satisfy the long-term requirements.
\end{itemize}
For both datasets, we feed the sequential data stream into the BioSLAM training procedure as depicted in \cref{fig:pipeline}.
Please note that each data will be only fed into the system once, and BioSLAM will not save the copy of that data sample.
For \textit{Campus} dataset, the LiDAR inputs, day- and night- visual inputs are fed into the system one by one.
For \textit{City} dataset, we will only feed the continuous LiDAR inputs to the system.

\subsection{Performance Evaluation}
\label{sec:performance}
To evaluate the localization performance on the large-scale \textit{City} dataset and long-term \textit{Campus} dataset in incremental learning, we divide the trajectory into individual trajectory segments and feed them into different place recognition systems in a sequential manner.
We evaluate the online localization performance mainly through \textit{Weighted Recall (WR)} of top-6 retrievals over incremental training, which is defined by $\textit{WR}=\sum_{k=1}^6 \omega_k r_{k},~\omega_1 =0.5, \omega_k=0.1 {\rm ~for~} k\neq 1$, where $r_{k}~(1 \le k \le 6)$ denotes the recall@k. Recall@k is the proportion of matched references found in the top-k retrieval. Specifically, the query image is deemed correctly localized (matched) if at least one of the top $k$ retrieved reference images is within the predefined neighbor range (e.g. 3m) from the ground truth position.

\begin{figure*}[ht]
	\centering
	\subfloat[{\small  Weighted recall in different areas}]{
    \includegraphics[width=0.47\linewidth]{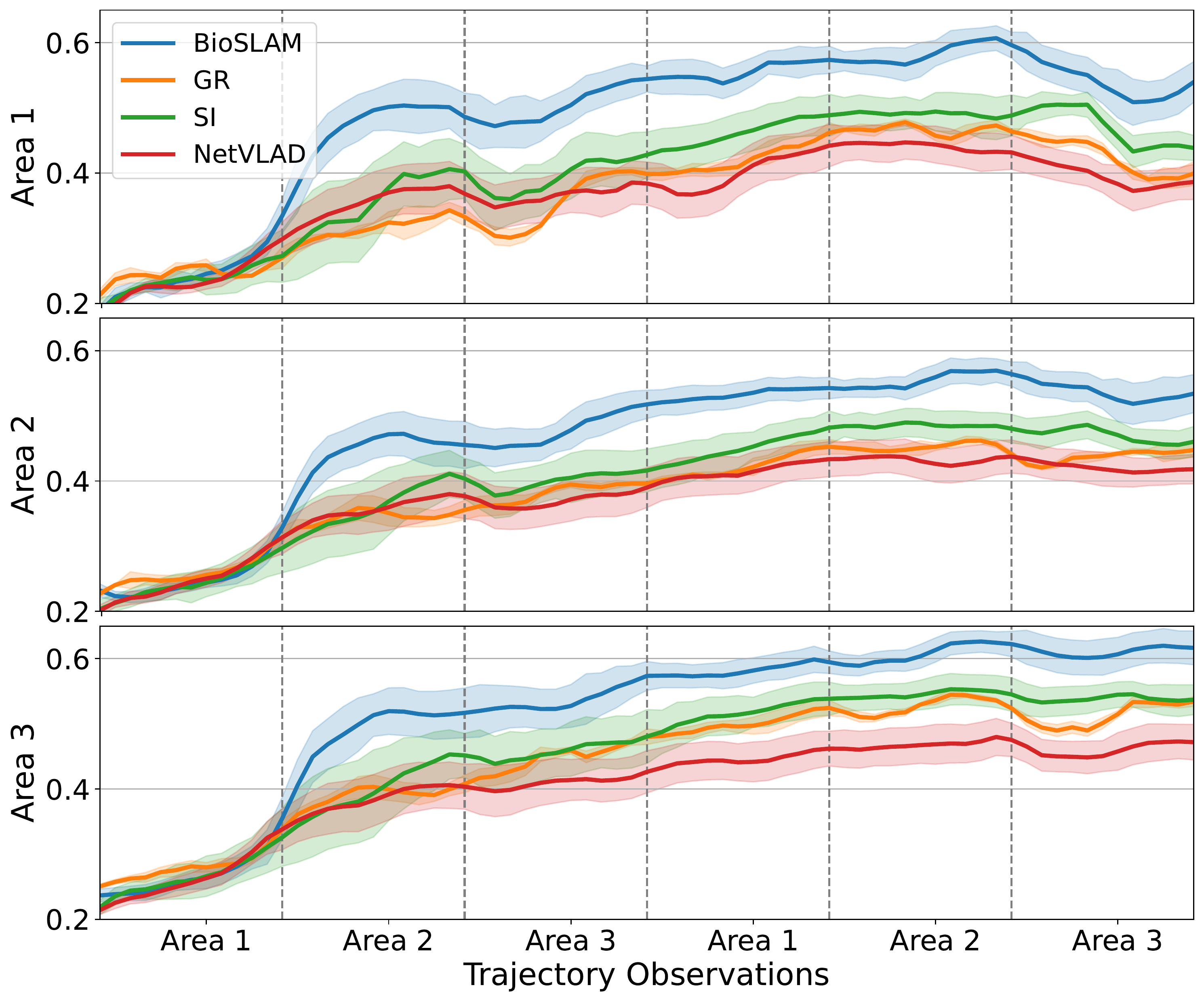}
     \label{fig:exp_pitt_lifelong_area}
     }
    	\subfloat[{\small Average weighted recall}]{
    \includegraphics[width=0.47\linewidth]{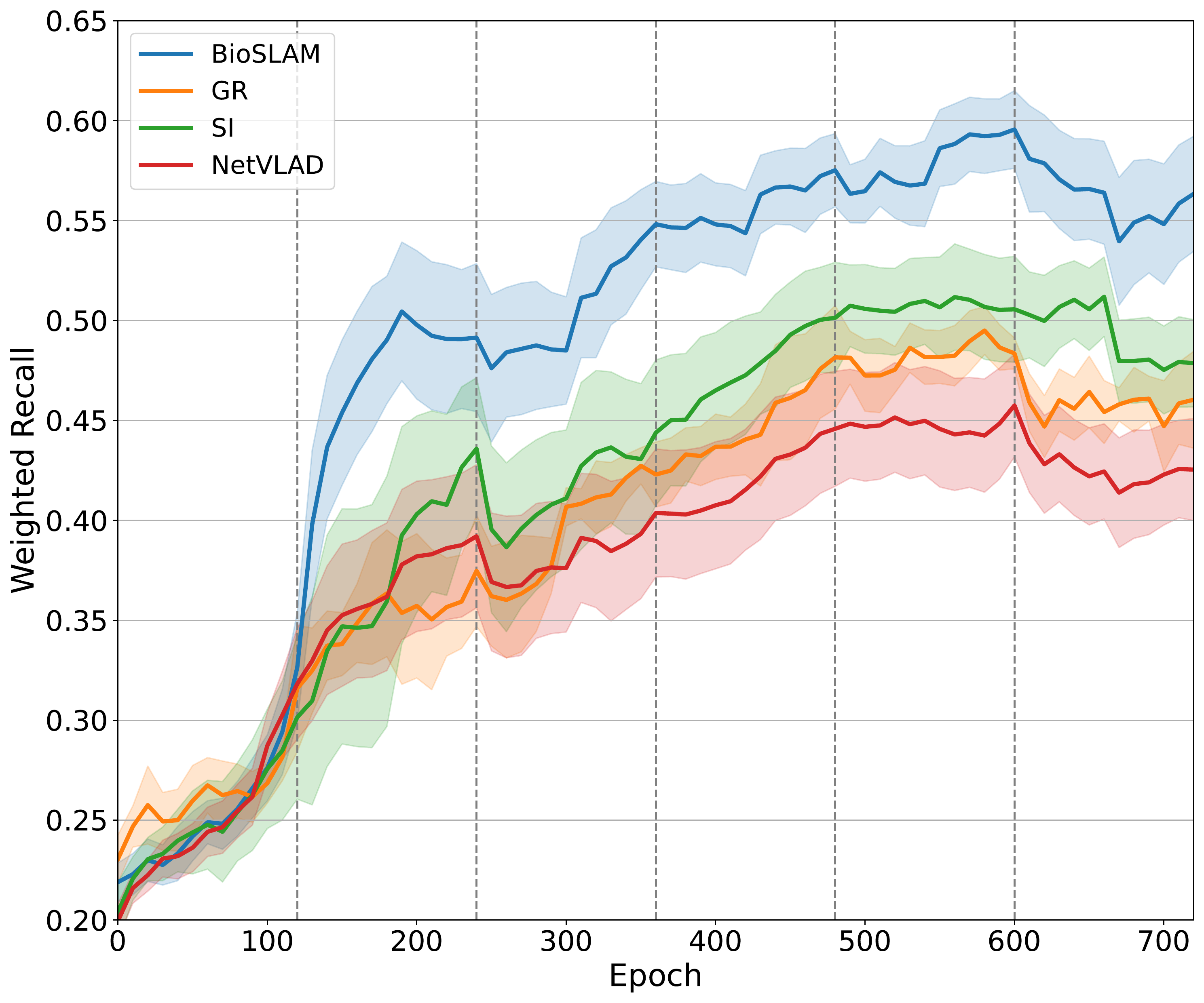}
    \label{fig:exp_pitt_lifelong_avg}
       }
    \vspace{-5pt}
	\caption{\textbf{Comparison of weighted recall w.r.t. training epochs on City dataset.} Place recognition methods incrementally trained on trajectory observations from 3 different areas. The shaded region shows the standard deviation.}
	\label{fig:exp_pitt_lifelong}
\end{figure*}

\subsection{Baselines}
Since our place recognition task involve different sensor modalities, non-learning/learning methods, and lifelong/non-lifelong methods, it is impossible to cover all the relevant state-of-the-arts.
We focus on the performance comparison from a 2D perspective and ignore Point-like~\cite{PR:pointnetvlad} 3D methods.
As a comparison, we select the following well-known non-learning methods (Bag-of-wards (BOW)~\cite{VPR:DBOW2}, CoHOG~\cite{VPR:CoHOG}), learning-based methods (NetVLAD~\cite{NetVLAD}, RegionVLAD~\cite{VPR:RegionVLAD}) and lifelong-based methods (Generative Replay (GR)~\cite{generative_play}, Synaptic Intelligence (SI)~\cite{SI}).
Among the above methods, GR and SI are the most related and important baselines to BioSLAM. Although  BioSLAM and GR both use memory replay, BioSLAM has more efficient and effective memory replay mechanisms. Because 1) BioSLAM replays samples according to reward (importance), while GR replays randomly and evenly. 2) BioSLAM has static memory to refresh the dynamic memory buffer to keep diverse and important memory traces, as well as easier to adapt to new trajectory observations. 

\color{black}

    \include{doc/exp_analysis}

\section{Experiment Analysis}
\label{sec:exp_analysis}
In this section, we analysis the lifelong place recognition results on both large-scale \textit{City} areas and long-term \textit{Campus} scenarios.
As shown in~\cite{lifelong_survey}, generative replay shows superior performance to other continual learning approaches.
The training quality is highly related to how the generated samples can better represent the entire data distribution. 
In our BioSLAM system, such ability is determined by the generative memory replay as stated in \cref{sec:generative_replay} and our BiLM system as stated in \cref{sec:BiLM}.
Specifically, we also investigate how different methods can handle the geometric and domain changes and how our BioSLAM system can achieve long-lasting memorization based on lifelong memory systems.

\begin{figure}[ht]
\centering
    \includegraphics[width=\linewidth]{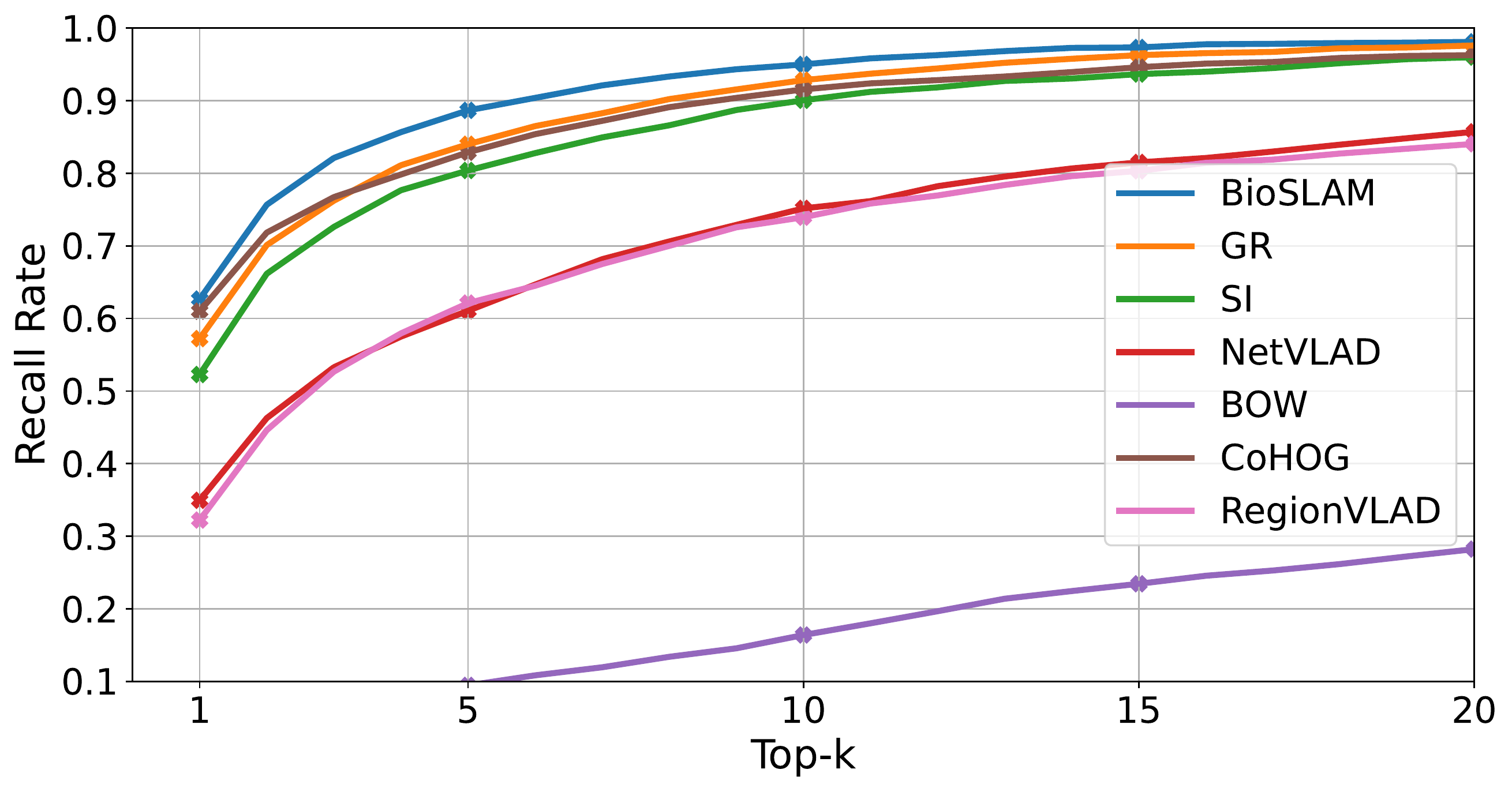}
    \vspace{-5pt}
    \caption{\textbf{Comparison of BioSLAM and baselines in terms of recall@k on City dataset.}}
    \label{fig:exp_pitt_recall_k}
\end{figure}

\subsection{Large-scale City Place Recognition}
\label{sec:exp_city}

We evaluate the performance of the BioSLAM in a large scale lifelong learning scenario with the City dataset. For the localization task under city-scale environments, robots may encounter multiple types of 3D geometric structures within the urban environments, such as open-street, bridges, parks, big buildings, and residential areas.
We divide  the $50$ trajectories with $120$km distance within the city into $3$ different areas based on their geometric properties: area 1 for commercial buildings, area 2 for parks, and area 3 for residential districts. For the large-scale City dataset, observations from different areas can be treated as different domains $D_t$ in \cref{eq:problem_define}.

In the training procedure, we incrementally feed the place recognition methods with trajectory observations from 3 different areas. 
\Cref{fig:exp_pitt_lifelong_area} shows the weighted recall curve of trajectory observations within area~1, area~2, and area~3, respectively. \Cref{fig:exp_pitt_lifelong_avg} shows the average weighted recall curve of all trajectories during training. 
As can be seen, BioSLAM outperforms other methods during training and is at least $14\%$ better than other baselines in terms of final average recall. 
More importantly, BioSLAM could keep the knowledge about previous trajectory observations when trained with new trajectory observations. For example, at epoch 240, when the training observations switched from area~2 to area~3, the performance drop on previous trajectories for BioSLAM is much smaller than in other methods, as shown in \cref{fig:exp_pitt_lifelong_avg}. Because BioSLAM retrains important previous knowledge by replaying related memory traces. Note that, BioSLAM replays important and highly rewarded memory traces, while GR only replays randomly. Thus BioSLAM has a much higher convergence rate and final performance than other baselines. 

After training, we evaluate the generalization ability of the final trained model on the fixed test set. \Cref{fig:exp_pitt_recall_k} shows the comparison between BioSLAM and other baselines in terms of top-$k$ recall on the test set of City dataset. As can be seen, although BioSLAM learns incrementally, it still performs better than baselines on classic (non-lifelong learning) offline test set evaluation, while some non-lifelong learning methods are designed or trained for offline evaluation.

    \subsection{Cross-domain Campus Place Recognition}
\label{sec:exp_campus}

\begin{figure*}[ht]
	\centering
	\subfloat[{\small Recalls per task (domain)}]{
    \includegraphics[width=0.47\linewidth]{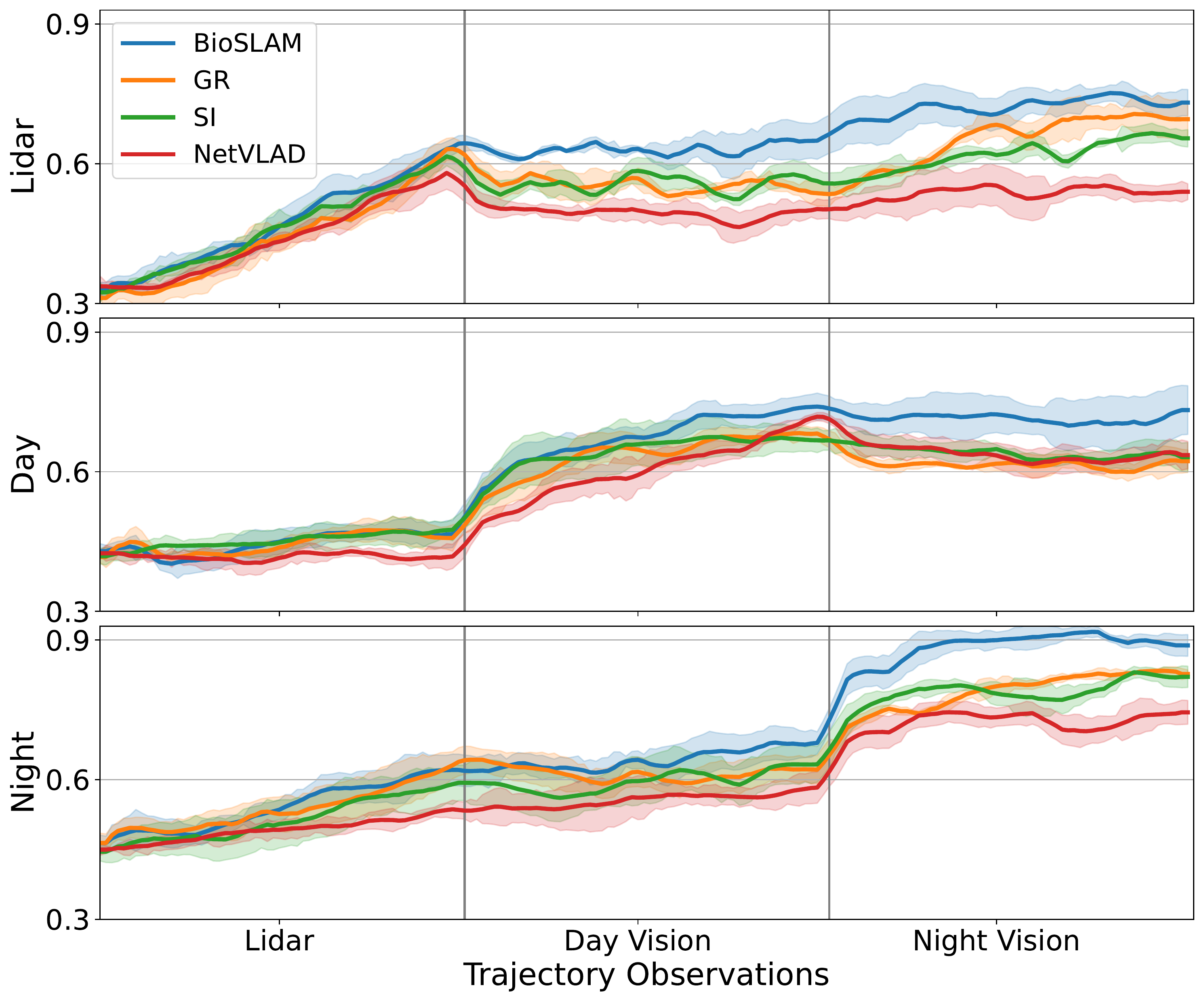}
    \label{fig:exp_cmu_lifelong_domain}
    }
    	\subfloat[{\small Average recall}]{
    \includegraphics[width=0.47\linewidth]{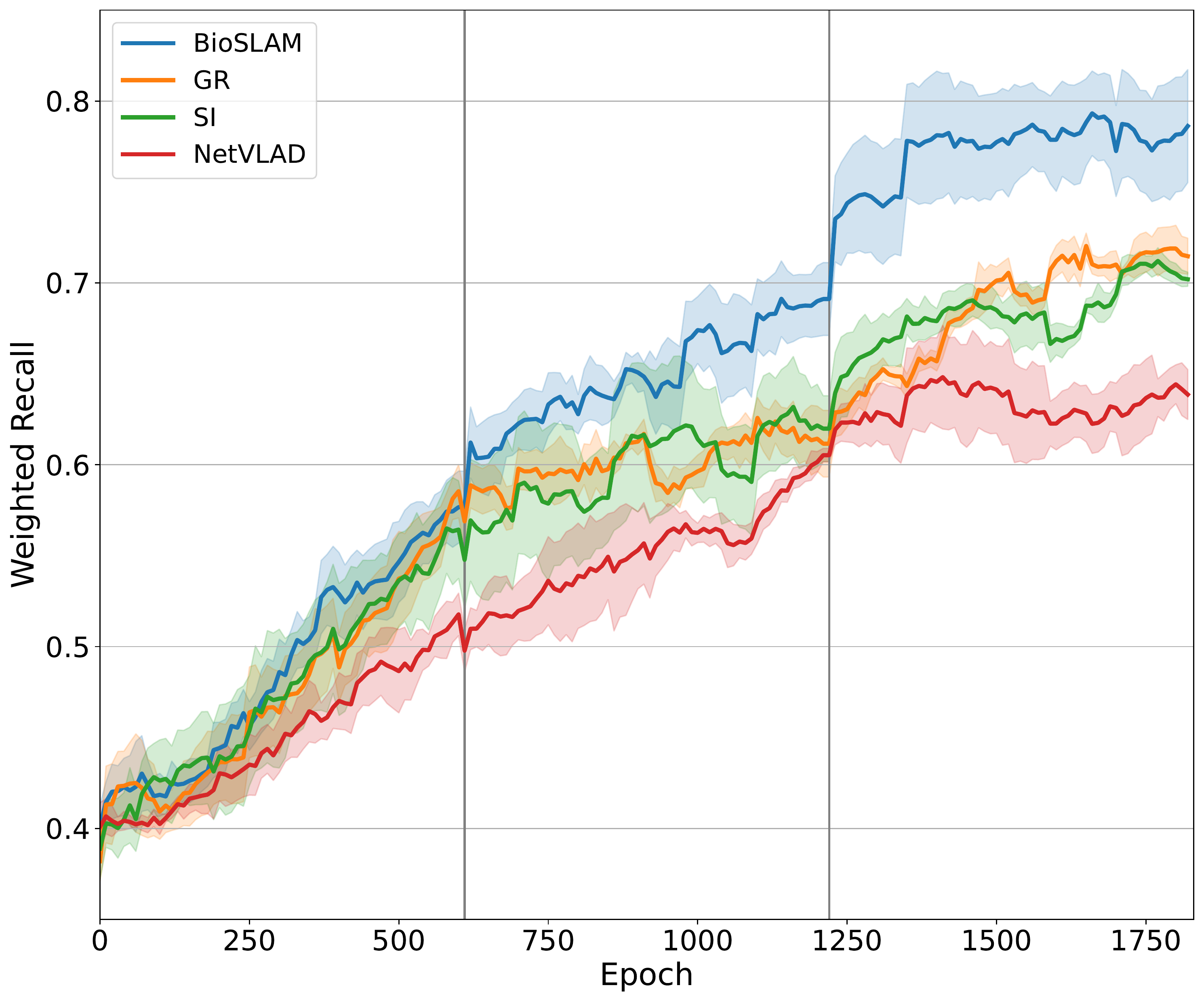}
    \label{fig:exp_cmu_lifelong_avg}
    }
    \vspace{-5pt}
	\caption{\textbf{Comparison of weighted recall w.r.t. training epochs on Campus dataset.}  Place recognition methods incrementally trained on trajectory observations from Lidar, day-time visual, and night-time visual inputs. The dashed region shows the standard deviation.
    }
	\label{fig:exp_cmu_lifelong}
\end{figure*}

\begin{figure}[ht]
\centering
    \includegraphics[width=\linewidth]{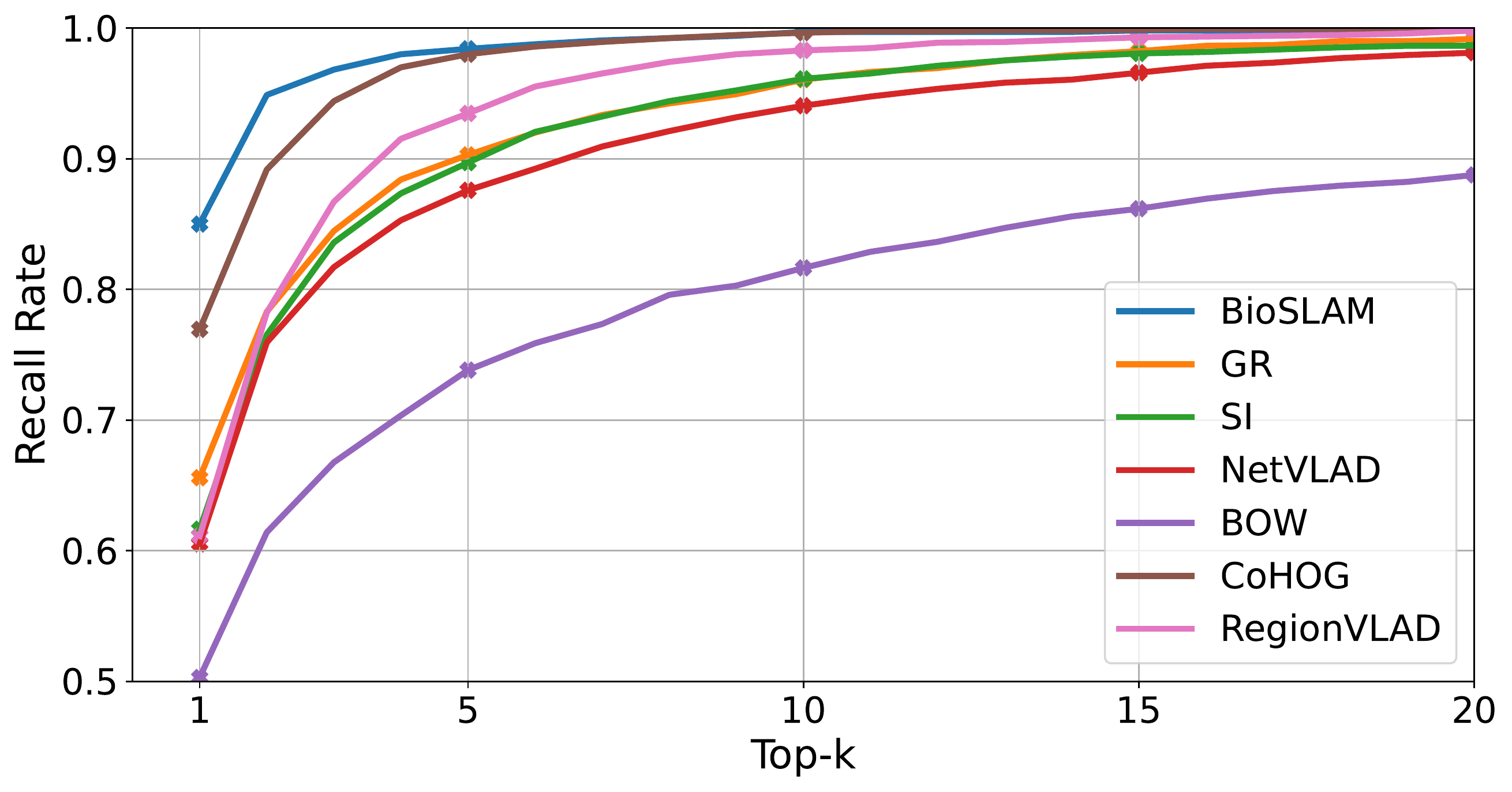}
    	\vspace{-5pt}
    \caption{\textbf{Comparison of BioSLAM and baselines in terms of recall@k on Campus dataset.}}
    \label{fig:exp_cmu_recall_k}
\end{figure}

Training place recognition models on independent domains are inefficient because no information will be shared. We thus demonstrate the merit of BioSLAM in more reasonable settings where the model benefits from solving place recognition from multiple domains (Lidar, day-time vision, night-time vision). A  place recognition model operating in multiple domains has several advantages. First, the knowledge of one domain can help better and faster understand other domains, because the domains are not completely independent in place recognition tasks. Second, generalization over multiple domains may result in more universal knowledge that applies to unseen domains. Such phenomenon is also observed in infants learning \cite{Infants_ability,bornstein2010development}. 

We evaluate the performance of BioSLAM on long-term and cross-domain lifelong learning scenario with Campus dataset. In the training procedure, we incrementally feed the place recognition methods with trajectory observations from different domains (ordered with Lidar, day-time vision, night-time vision), and evaluate the performance on all domains. Observations from Lidar, day-time vision, and  night-time visual signals can be treated as different domains $D_t$ in \cref{eq:problem_define}.

\begin{figure*}[th]
	\centering

	\subfloat[{\small Ablation study on City dataset}]{
    \includegraphics[width=0.47\linewidth]{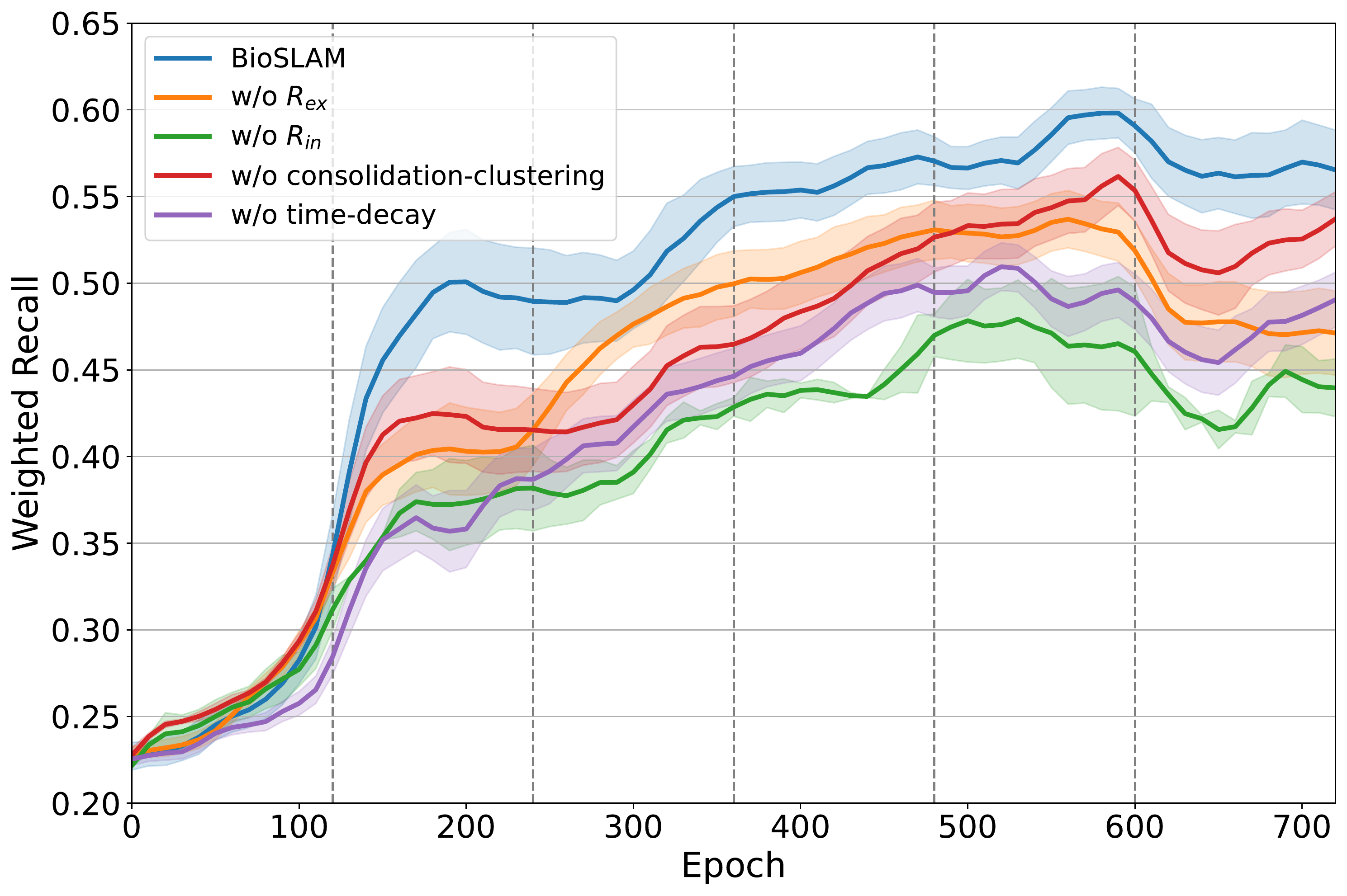}
    \label{fig:exp_pitt_ablation}
    }
    	\subfloat[{\small Ablation study on Campus dataset}]{
    \includegraphics[width=0.47\linewidth]{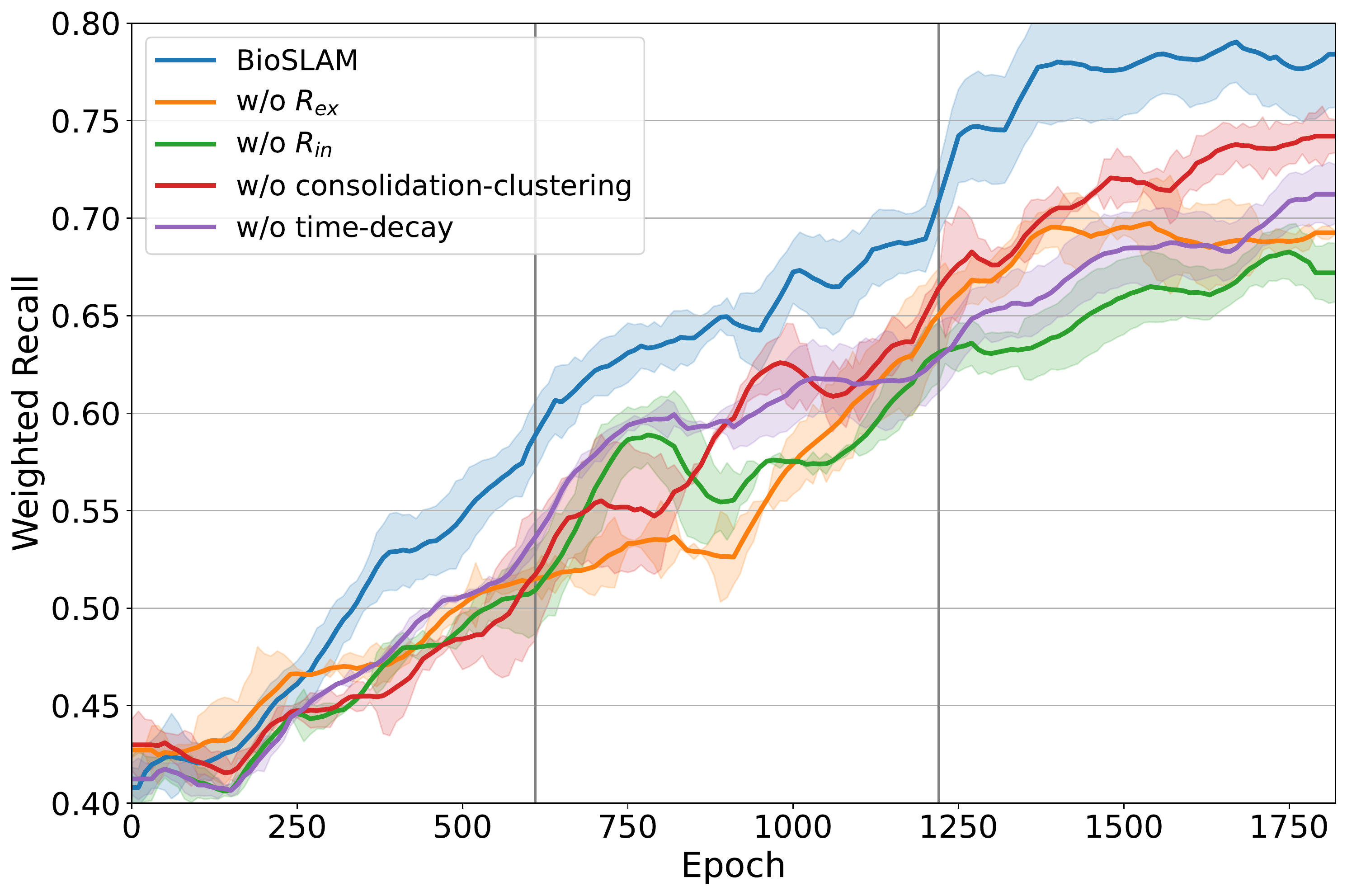}
    \label{fig:exp_cmu_ablation}
    }
    	\vspace{-5pt}
	\caption{\textbf{Ablation study of BioSLAM.} Comparison between BioSLAM and its variants: (1) w/o $\mathcal{R}_{ex}$, (2)  w/o $\mathcal{R}_{in}$, (3)  w/o clustering in memory consolidation, (4) w/o time-decay in memory refresh. 
    }
	\label{fig:exp_ablation}
\end{figure*}

\Cref{fig:exp_cmu_lifelong_domain} shows the performance comparison between BioSLAM and baselines on different domains of the Campus dataset. In $0 \sim 600$ epochs, we train the place recognition model on the Lidar domain and the performance of all methods on all domains increases within 600 epochs. This verifies that the knowledge of one domain can help better and faster understand other domains. 
In $600 \sim 1200$ epochs, we train the model in the day-time visual domain. For all methods, the performance in the day-time visual domain increases but the Lidar performance decrease around the switching point of epoch=600. This is reasonable because the background of Lidar and day-time visual images are totally different. In the Lidar domain, the performance drop of BioSLAM is much smaller than in other methods. This shows that BioSLAM could learn observations in a new domain without forgetting observations from past domains.
In $1200 \sim 1800$ epochs, we train the model in the night-time visual domain.
In addition to the performance increases in the night-time visual domain, the performance of BioSLAM also increases in the Lidar domain. With efficient replay mechanisms of  BioSLAM, the knowledge of one domain can help better understand other domains.    
The average weighted recall of all domains is shown in \cref{fig:exp_cmu_lifelong_avg}. As can be seen, the performance of all methods is similar in the beginning, but as new observations from new domains are added, BioSLAM converges faster and better than baselines and outperforms other methods by at least 10\%  in terms of final average recall. 

After training, we evaluate the cross-domain generalization ability of the final trained model on the fixed test set. \Cref{fig:exp_cmu_recall_k} shows the comparison between BioSLAM and other baselines in terms of top-$k$ recall on the test set of the Campus dataset. As can be seen, although BioSLAM learns incrementally, it still performs better than baselines on classic (non-lifelong learning) offline test set evaluation, while some non-lifelong learning methods are designed or trained for offline evaluation.

\begin{table}[ht]
\centering
\caption{Comparison of weighted recall (\%) on city and campus datasets.}
\begin{tabular}{ll|cc}
\hline \hline
\multicolumn{2}{c|}{Weighted Recall (\%)}                                                                                        & City    & Campus        \\ \hline
\multicolumn{1}{l|}{\multirow{2}{*}{Non-learning}}                                                            & BOW              & 5.7           & 60.1          \\
\multicolumn{1}{l|}{}                                                                                         & CoHOG            & 70.1          & 85.1          \\ \hline
\multicolumn{1}{l|}{\multirow{2}{*}{\begin{tabular}[c]{@{}l@{}}Learning based\\ (not lifelong)\end{tabular}}} & RegionVLAD       & 45.3          & 75.1          \\
\multicolumn{1}{l|}{}                                                                                         & NetVLAD          & 47.8          & 72.2          \\ \hline
\multicolumn{1}{l|}{\multirow{3}{*}{Lifelong learning}}                                                       & SI               & 65.1          & 73.7          \\
\multicolumn{1}{l|}{}                                                                                         & GR               & 68.4          & 76.1          \\
\multicolumn{1}{l|}{}                                                                                         & \textbf{BioSLAM} & \textbf{73.6} & \textbf{91.2} \\ \hline \hline
\end{tabular} 
\label{tab:exp_recall_topk}
\end{table}

\Cref{tab:exp_recall_topk} shows the comparison between different methods both on the fixed test set of City and Campus datasets. In City dataset, BioSLAM outperforms a state-of-the-art lifelong learning method GR by 7.6 \%, and a non-learning method CoHOG by 5\%.  In Campus dataset, BioSLAM outperforms a lifelong learning method GR by 19.8 \%, and a non-learning method CoHOG by 7.2\%. Note that, this paper focuses on incremental and lifelong learning scenarios, so the most important evaluation metric is the recall curve over incremental learning (as shown in \cref{fig:exp_pitt_lifelong,fig:exp_cmu_lifelong}). For a recall on the fixed test set, some non-lifelong learning methods (i.e. CoHOG) may perform very well, but these methods can not learn incrementally, so the performance of the non-lifelong learning methods is limited. Thus lifelong learning methods have a higher potential in wider and changing real environments.

\subsection{Ablation Study}
\label{sec:ablation}

As mentioned in \cref{sec:BiLM}, BioSLAM has several novel mechanisms that differ from previous lifelong learning methods: (1) external reward $\mathcal{R}_{ex}$ to indicate localization performance; (2) internal reward $\mathcal{R}_{in}$  to indicate the robustness of feature representation; (3) Static memory consolidation to abstract concise memory traces, and clustering (\cref{eq:clusters}) is the key of memory consolidation; (4) Dynamic memory refreshing to effectively replay important memory, in which, the time-decay mechanism (\cref{eq:mem_refresh}) for importance weight is critical. 
We further evaluate the effectiveness of the above  mechanisms of BioSLAM by comparing BioSLAM with the following variants: (1) w/o $R_{ex}$: without applying external reward, then \cref{eq:reward_total} becomes $\mathcal{R}_k =\mathcal{R}_{in}(q_k)$; (2) w/o $R_{in}$: without applying internal reward, then  \cref{eq:reward_total} becomes $\mathcal{R}_k =\mathcal{R}_{ex}(q_k)$; (3) w/o
consolidation-clustering: without using clustering in static memory consolidation, then \cref{algo:consolidation} becomes directly storing all memory traces in static memory; (4) w/o time-decay: without using the time decay factor in importance sampling, which is equivalent to set $\gamma=1$ in dynamic memory refresh \cref{eq:mem_refresh}. Note that these variants follow the control variates method. They cover all important mechanisms of BioSLAM without overlapping functionalities.

The results for the ablation study on City dataset are shown in \cref{fig:exp_pitt_ablation}, and the results on Campus dataset are shown in \cref{fig:exp_cmu_ablation}. As can be seen, BioSLAM outperforms its variants on both City and Campus datasets. The removal of any component leads to a significant performance drop. 
In particular, the performance grain of BioSLAM with respect to “w/o time-decay" validates the necessity of decay weights (rewards) of importance sampling during dynamic memory refresh. 
The larger performance drop is caused by removing internal rewards, which means the indicator (internal reward) for the robustness of feature representation is critical to retrieving memories. The performance of “ w/o
cluster-consolidation" is close to BioSLAM, but BioSLAM is much memory efficient by clustering and downsampling.

\begin{figure*}[thbp]
 \hspace{-10pt}
    \subfloat[{\small Similarity matrix from different areas on City dataset}]{
    \includegraphics[width=0.495\linewidth]{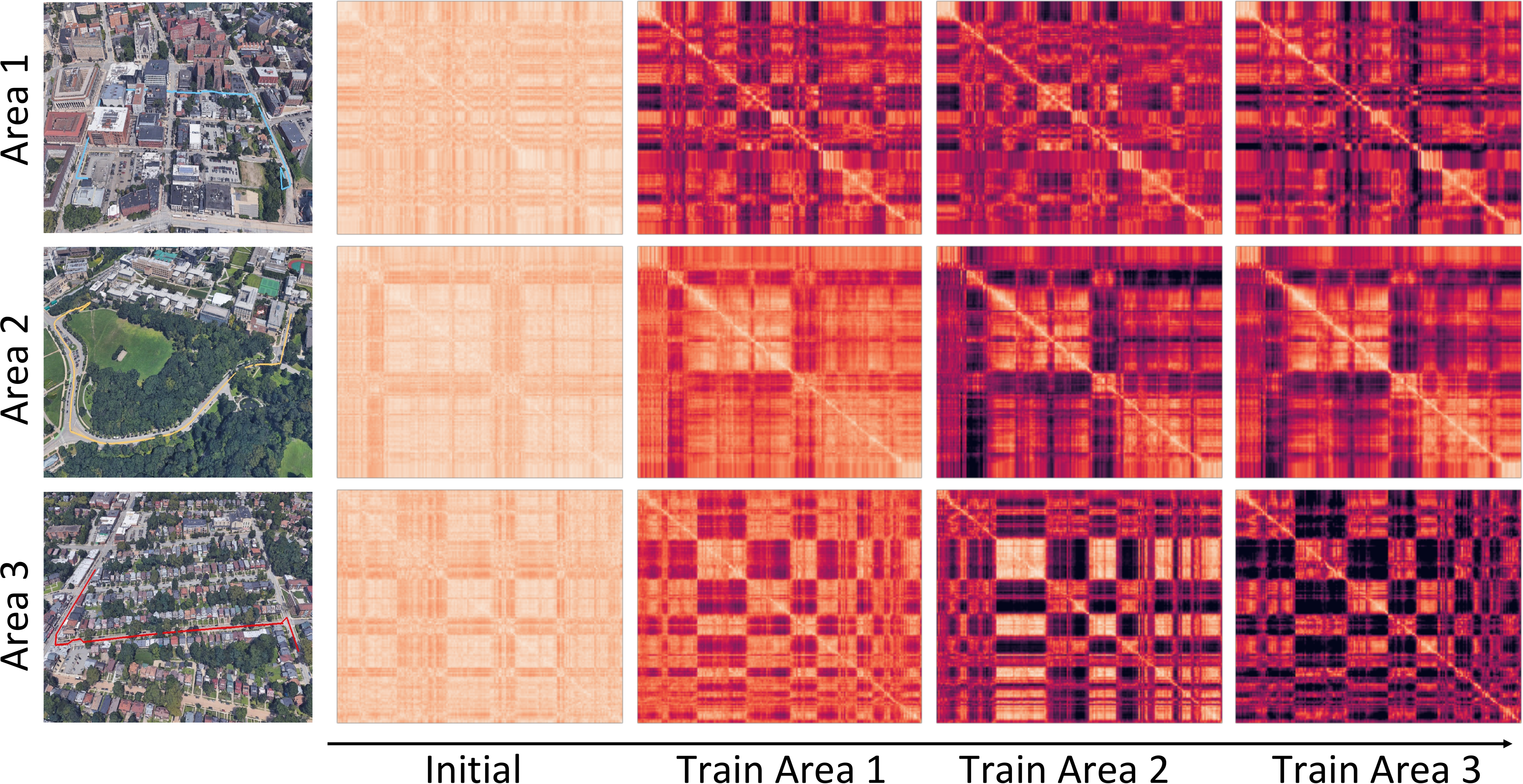}
    \label{fig:exp_pitt_diff}
    }
    \hspace{-10pt}
    \subfloat[{\small Similarity matrix from different domains on Campus dataset}]{
    \includegraphics[width=0.51\linewidth]{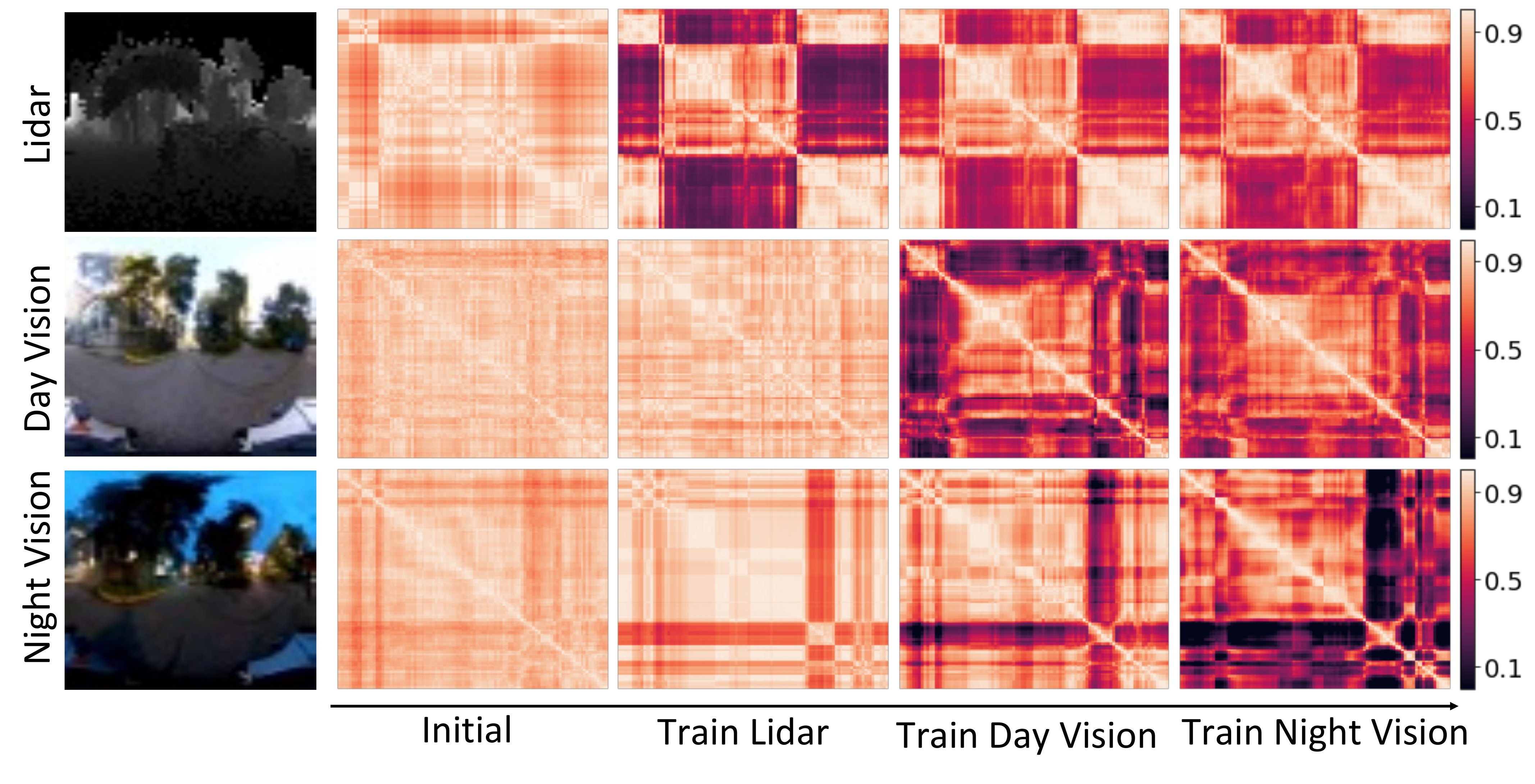}
    \label{fig:exp_cmu_diff}
    }
    \vspace{-5pt}
    \caption{\textbf{Similarity matrix over training}. (a) City dataset. The left column represents the trajectories within area 1, area 2, and area 3. The right three columns represent the corresponding similarity matrices over training. (b) Campus dataset. The left column represents  Lidar, day-time visual, and night-time visual observations. The right three columns represent the  corresponding similarity matrices over training. }
\end{figure*}



\begin{figure*}[thbp]
 \hspace{-7pt}
    \subfloat[{\small PCA visualization on City dataset}]{
    \includegraphics[width=0.501\linewidth]{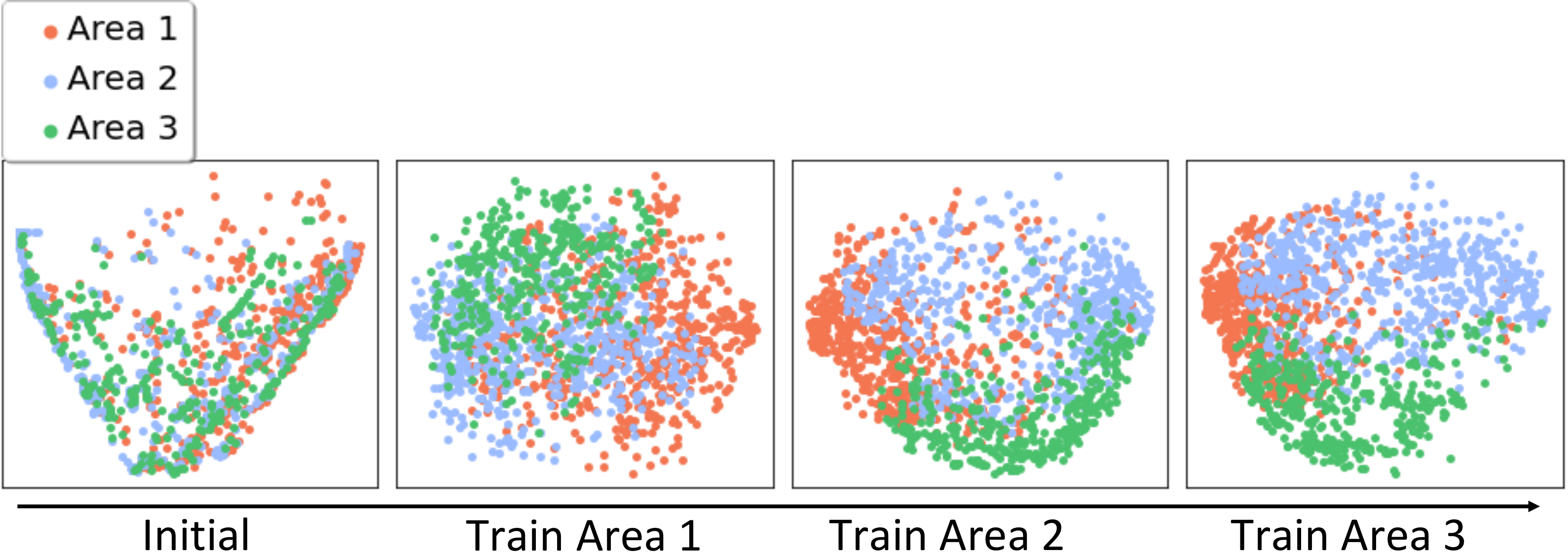}
    \label{fig:pitt_pca}
    }
    \hspace{-9pt}
    \subfloat[{\small PCA visualization on Campus dataset}]{
    \includegraphics[width=0.501\linewidth]{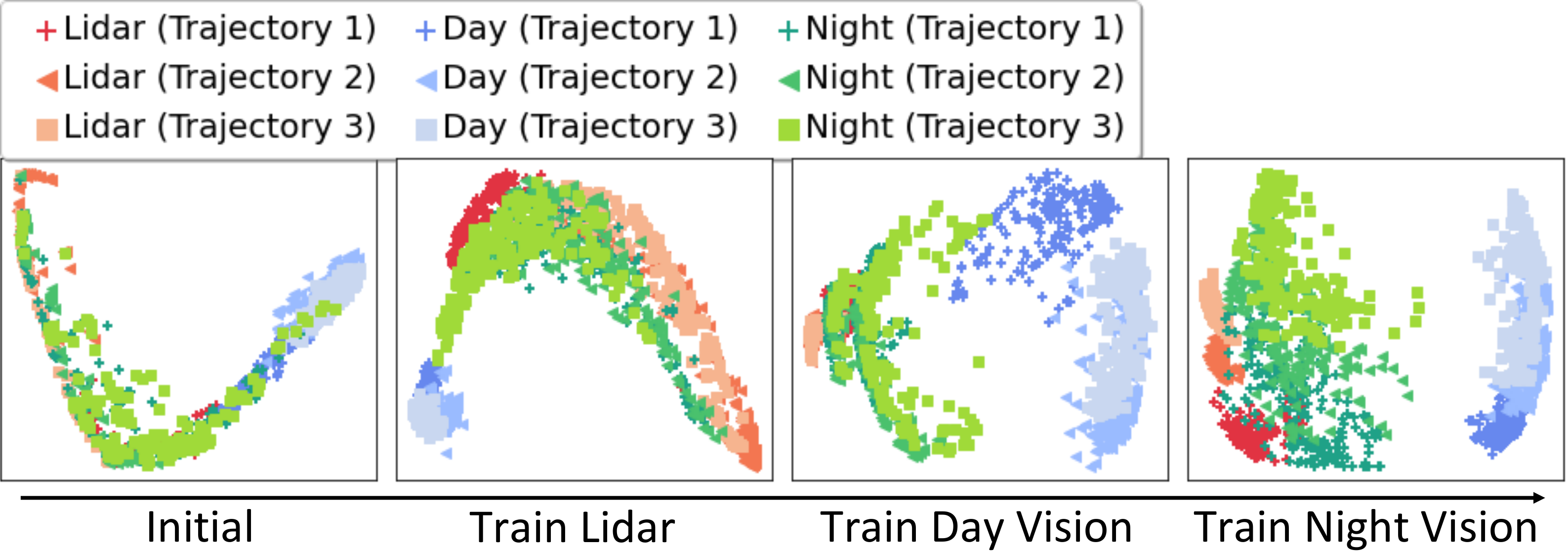}
    \label{fig:cmu_pca}
    }
    \vspace{-5pt}
    \caption{\textbf{PCA visualization over training}. (a) City dataset. Visualization of observations from different areas with PCA.  (b) Campus dataset. Visualization of observations from different trajectories and different domains with PCA.  }
\end{figure*}

\subsection{BioSLAM Feature Property}
\label{Sec:LFP}

In this section, we visualize and evaluate the BioSLAM learned features (place descriptor, \cref{eq:lean_feature}) $\mathcal{F}(q_k)$ with similarity matrix and Principle Component Analysis (PCA). Similarity Matrix $M_{\text{sim}}$ is defined by the cosine similarity between reference $O^D$ and query $Q^D$ features, with  $M_{\text{sim}}(i,j)=\cos\left(\mathcal{F}(O_i^D), \mathcal{F}(Q_j^D)\right)$. A high-contrast similarity matrix indicates that the learned feature $\mathcal{F}$ has a strong expression and discrimination ability.

The similarity matrix of BioSLAM over training on City dataset is shown in \cref{fig:exp_pitt_diff}.
The left column represents the sampled trajectories from area 1, area 2, and area 3. The right three columns represent the similarity matrices of the corresponding trajectories  (from left to right) after incrementally training on area 1, area 2, and area 3, respectively.
After training on area 1, the similarity matrices of all areas increase contrast. Then training on  area 2 and area 3, the similarity matrix of area 1 is almost non-decayed. 
That means BioSLAM still has strong expression ability on past trajectories when learning from different areas. 

The similarity matrix of BioSLAM over training on Campus dataset is shown in \cref{fig:exp_cmu_diff}. The left column represents the sampled observations from Lidar, day-time visual, and night-time visual inputs. The right three columns represent the similarity matrices of the corresponding observations  (from left to right) after incrementally training on Lidar, day-time visual, and night-time visual domains, respectively.
After training on the Lidar domain, the similarity matrix of the Lidar observations increases contrast. Then incrementally training the model on day-time and night-time visual domain, the similarity matrix of the corresponding domain become more contrastive, and the similarity matrix of Lidar almost does not decay. That means BioSLAM could remember the past domains when learning from totally different domains. 

We used PCA to reduce the dimension of BioSLAM learned features to 2d. The PCA visualization of learned features of observations from different areas on the City dataset is shown in \cref{fig:pitt_pca}. The subfigures from left to right represent PCA visualization results at the initial step, and after incrementally training on area 1, area 2, and area 3.
As can be seen, with BioSLAM training, observations within the same area are almost clustered together. The clusters of the different areas become easier to discriminate over incremental learning. 

The PCA visualization of learned features from different domains and trajectories on the Campus dataset is shown in \cref{fig:cmu_pca} . (For clearer visualization, we only visualize three trajectory segments in each domain).  
The sub-figures from left to right represents PCA visualization results at the initial step, and after incrementally training on Lidar, day-time visual, and night-time visual domains. As can be seen, BioSLAM not only differentiates different domains (3 clusters from left to right) but also differentiates different trajectories within each domain. As shown in the right sub-figure in \cref{fig:cmu_pca}, the same trajectories of different domains are relatively close. 
As an example, for trajectory 1,  the PCA results of the Lidar domain and night-time visual domains are close to each other and located in the lower part of the PCA visualization results. That means, BioSLAM could incrementally learn trajectories from different domains, and it has the potential to find the cross-domain relationship of place observations from different domains.

\color{black}
\subsection{BioSLAM Memory Activity}
\label{Sec:LMA}

\begin{figure*}[thbp]
\hspace{-5pt}
    \subfloat[{\small Proportion of different trajectories in dynamic memory}]{
    \includegraphics[width=0.43\linewidth]{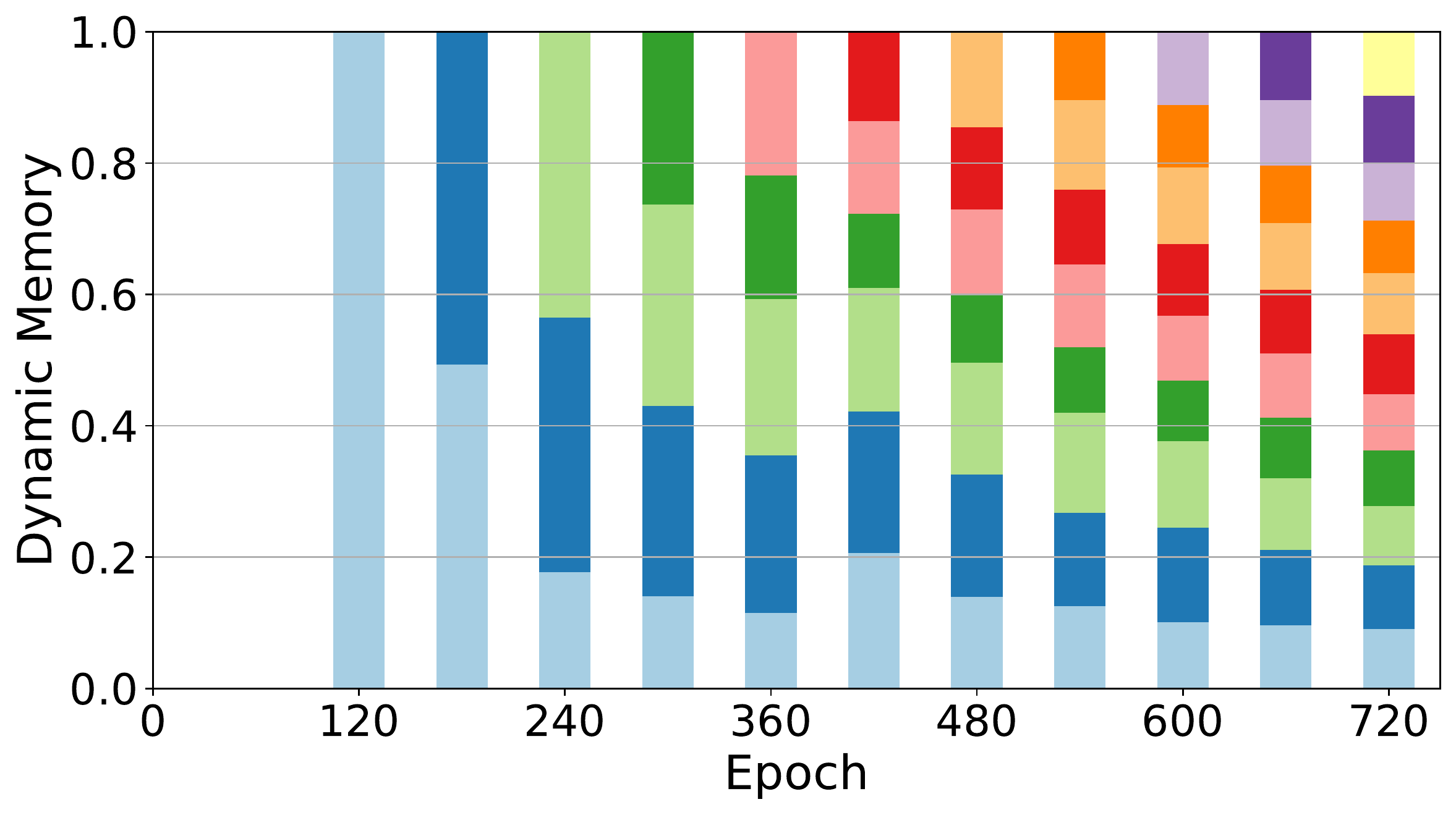}
    \label{fig:pitt_memory}
    }
    \hspace{-5pt}
    \subfloat[{\small Reward ratio of different trajectories}]{
    \includegraphics[width=0.55\linewidth]{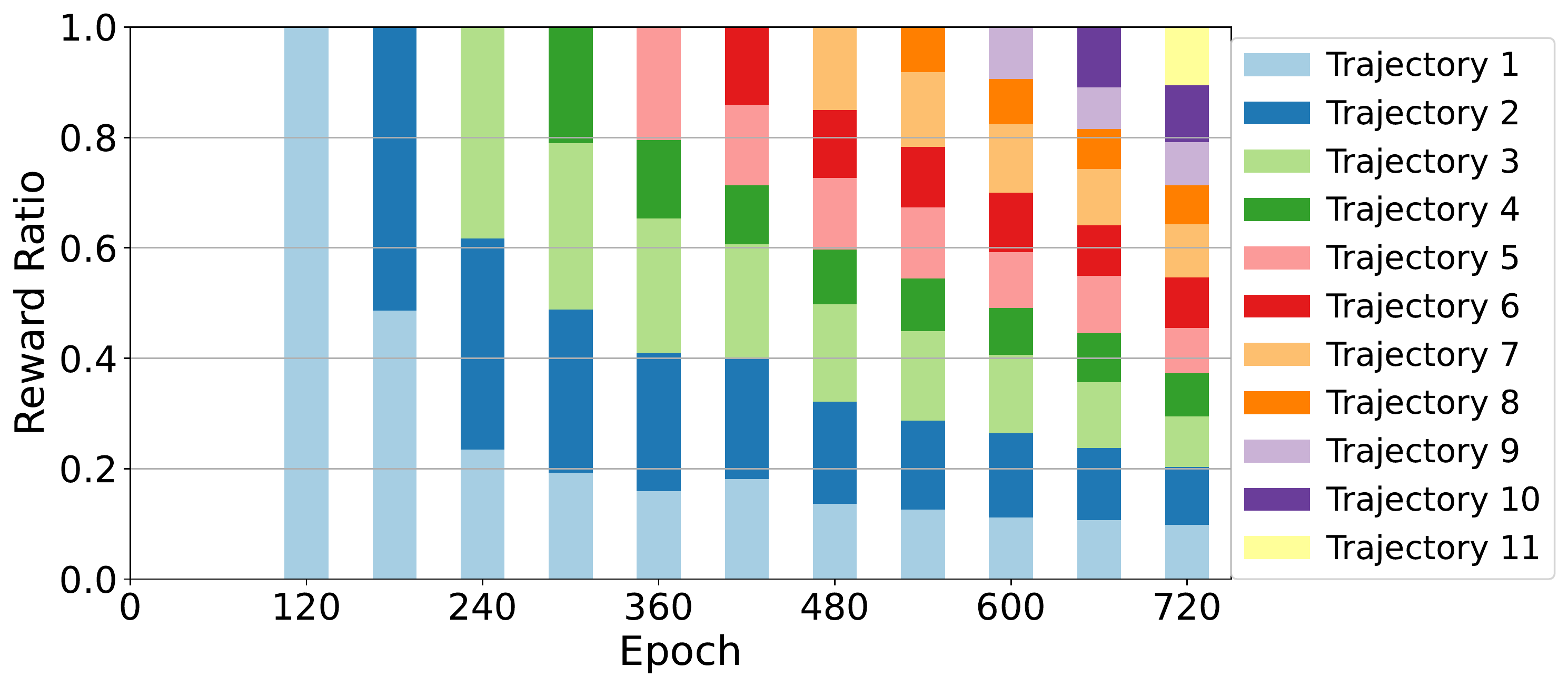}
    \label{fig:pitt_reward}
    }
    \vspace{-5pt}
    \caption{\textbf{Dynamic memory zone and Reward ratio (normalized) of different trajectories over training on City dataset.}}
\end{figure*}

\begin{figure*}[thbp]
 \hspace{-5pt}
    \subfloat[{\small  Proportion of different domains in dynamic memory}]{
    \includegraphics[width=0.43\linewidth]{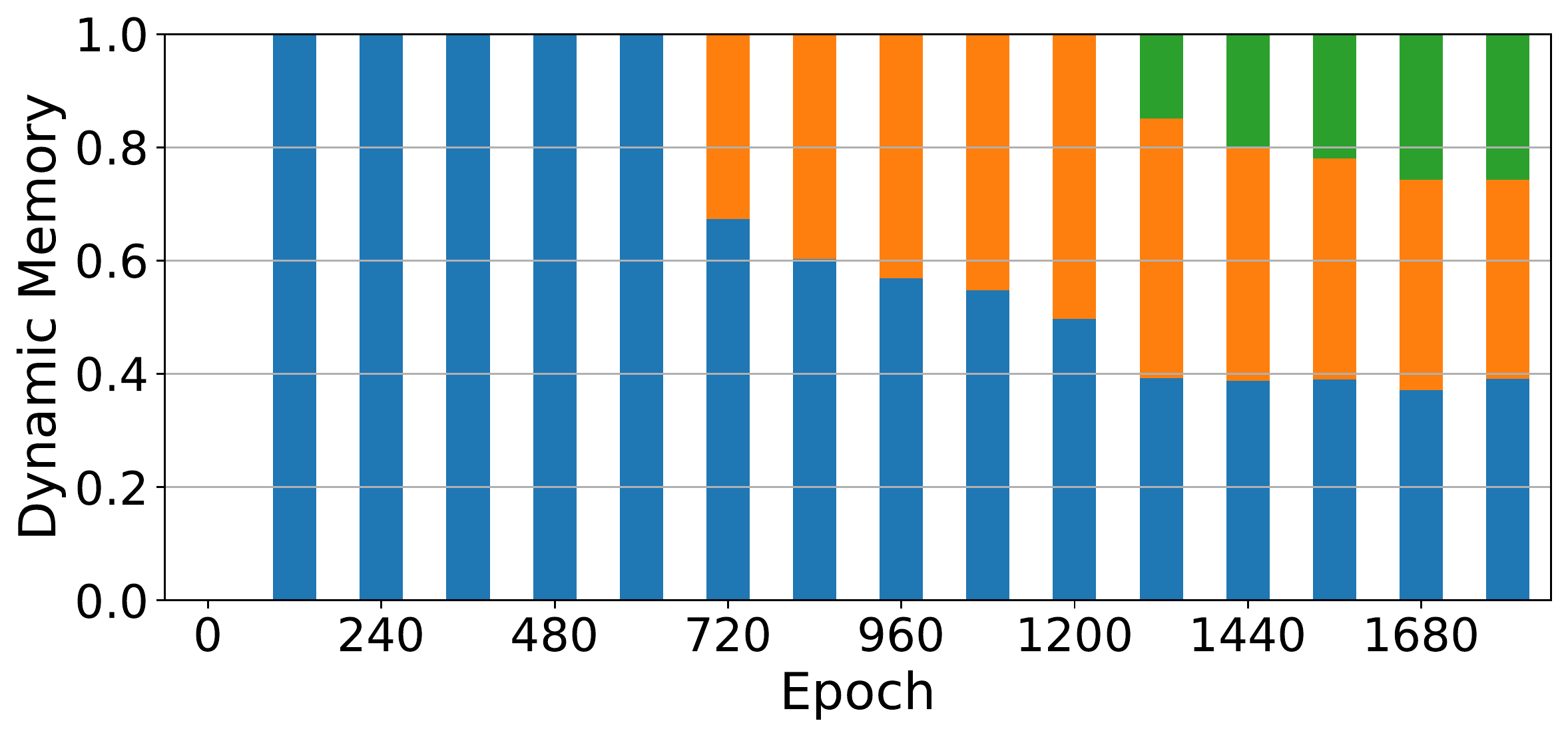}
    \label{fig:cmu_memory}
    }
    \hspace{-5pt}
    \subfloat[{\small Reward ratio of different domains}]{
    \includegraphics[width=0.55\linewidth]{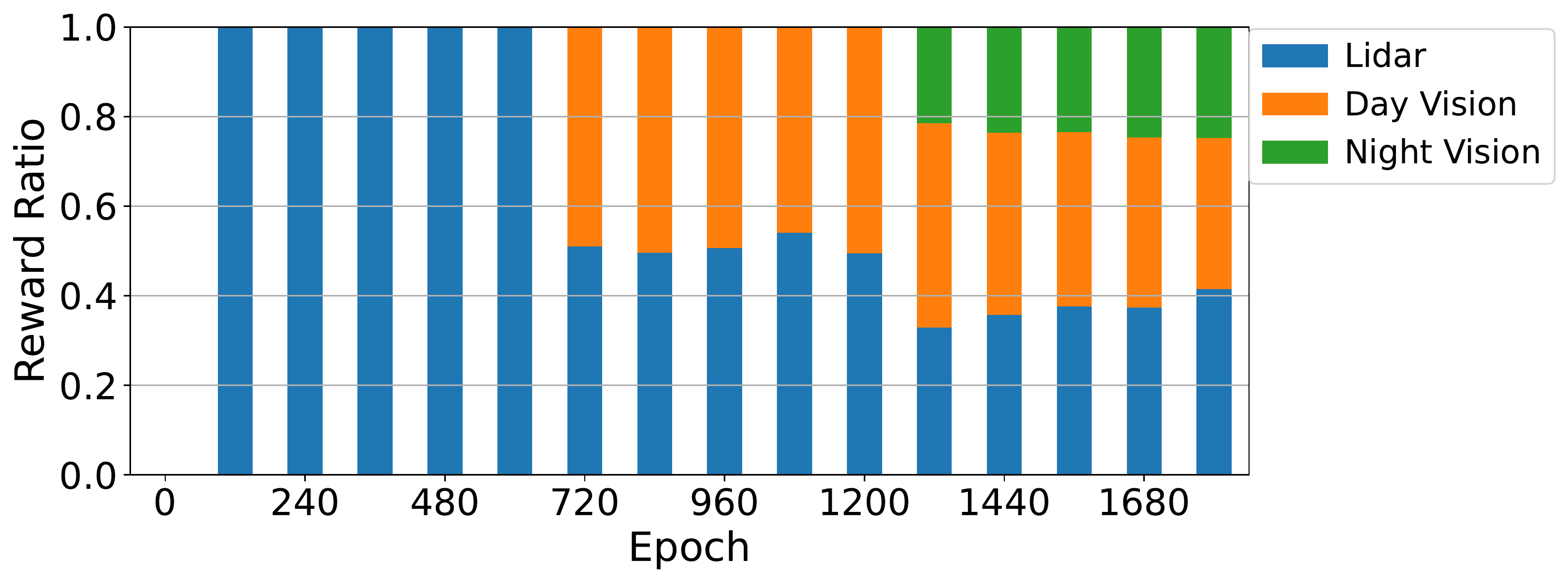}
    \label{fig:cmu_reward}
    }
    \vspace{-5pt}
    \caption{\textbf{Dynamic memory zone and Reward ratio (normalized) of different domains over training on Campus dataset.}}
\end{figure*}

As described in \cref{sec:memory}, static memory $M_s$ stores selected memory traces by memory consolidation. Because clustering and downsampling are based on feature and spatial property (\cref{algo:consolidation}), the static memory stores concise and diverse observations in terms of feature and spatial property. Then dynamic memory $M_d$ samples memory traces from static memory with importance sampling, and the sampling weight is proportional to its reward value. 

The memory traces in the dynamic memory zone has a direct impact on lifelong learning performance because the replayed samples from dynamic memory are used for training at every iteration.  In this section, we visualize the dynamic memory zone to see the proportion of memory traces from different domains or trajectory segments. 
Note that, if memory traces of a domain have higher rewards than other domains, then the dynamic memory zone holds more memory traces (samples) from the domain. 
Thus, to better understand the proportion between different domains, we also visualize the (normalized) reward ratio of each domain or trajectory segment. For trajectory $i$, $\textit{reward\_ratio}(i)=\frac{ \bar{R}_{(i)}}{\sum_j \bar{R}_{(j)}}$, where $\bar{R}_{(i)}$ is the  average reward for observations from trajectory $i$.  

For the City dataset, the proportion of observations from different trajectory segments in the dynamic memory zone is shown in \cref{fig:pitt_memory}.  As the new trajectory segments incrementally feed into BioSLAM (every 60 epochs), the trajectory diversity in the dynamic memory zone increases. \Cref{fig:pitt_reward} shows the reward ratio of different trajectories of training. As can be seen, the proportion of different trajectories in the dynamic buffer is consistent with the reward ratio of corresponding trajectories. Because given a trajectory segment, a higher reward means worse performance, BioSLAM uses higher sampling weights  to retrieve more memory traces from the high-rewarded trajectory to achieve better performance. 
As an example, from epoch 360 to 420, the reward ratio of trajectory 1 increases in \cref{fig:pitt_reward}, and dynamic memory samples more memory traces of trajectory 1.

For the Campus dataset, the proportion of observations from different domains in the dynamic memory zone is shown in \cref{fig:cmu_memory}. \Cref{fig:cmu_reward} shows the reward ratio of different domains of training. As can be seen, the proportion of different domains in the dynamic buffer is consistent with the reward ratio of corresponding domains. 
As an example, in the final step, the reward ratio of the night-visual domain is lower than other domains in \cref{fig:cmu_reward}, which means BioSLAM already achieves better performance in the night-visual domain. Then the dynamic memory tends to retain only a small amount of night-visual memory traces, leaving valuable memory zone for other high-rewarded domains (i.e. LiDAR).

Thus, we have the following relationship between dynamic memory zone and rewards: the performance on a domain (or trajectory) $i$ is lower $\rightarrow$ higher reward on the domain (or trajectory)  $i~\rightarrow$ more memory traces of domain (or trajectory) $i$ in dynamic memory $\rightarrow$ training on more replayed samples from the domain (or trajectory) $i~\rightarrow$ the performance on the domain (or trajectory) $i$ may increases. 
The above relationship between dynamic memory and rewards can serve as feedback compensation.\footnote{Note that, the performance on the domain (or trajectory)  $i$ may be saturated and not increase. But the relationship still works as feedback compensation.}

    \subsection{Incremental Confidence}
\label{sec:exp_confidence}

\begin{figure*}[thbp]
 \hspace{-10pt}
    \subfloat[{\small Confidence map from different areas on City dataset}]{
    \includegraphics[width=0.492\linewidth]{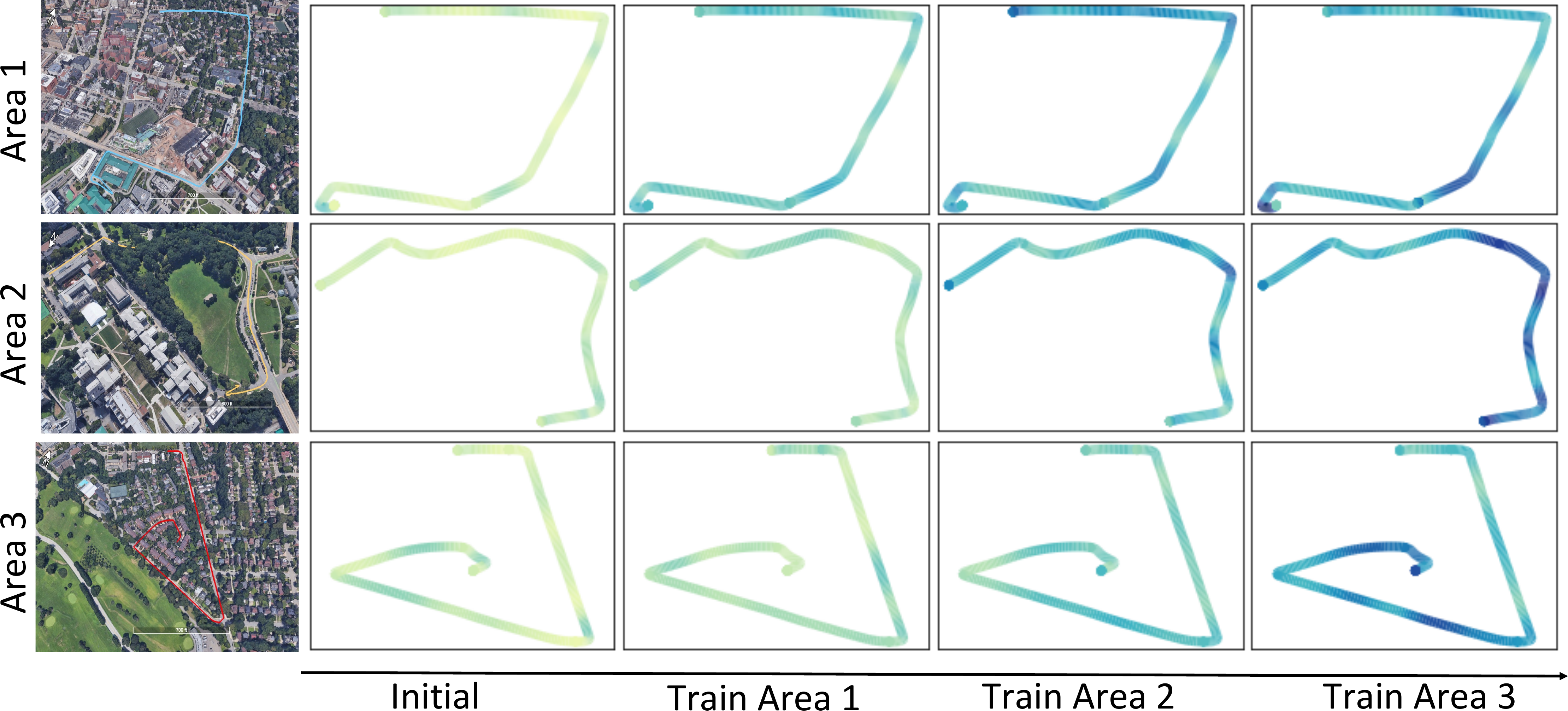}
    \label{fig:exp_pitt_confidence}
    }
    \hspace{-10pt}
    \subfloat[{\small Confidence map from different domains on Campus dataset}]{
    \includegraphics[width=0.51\linewidth]{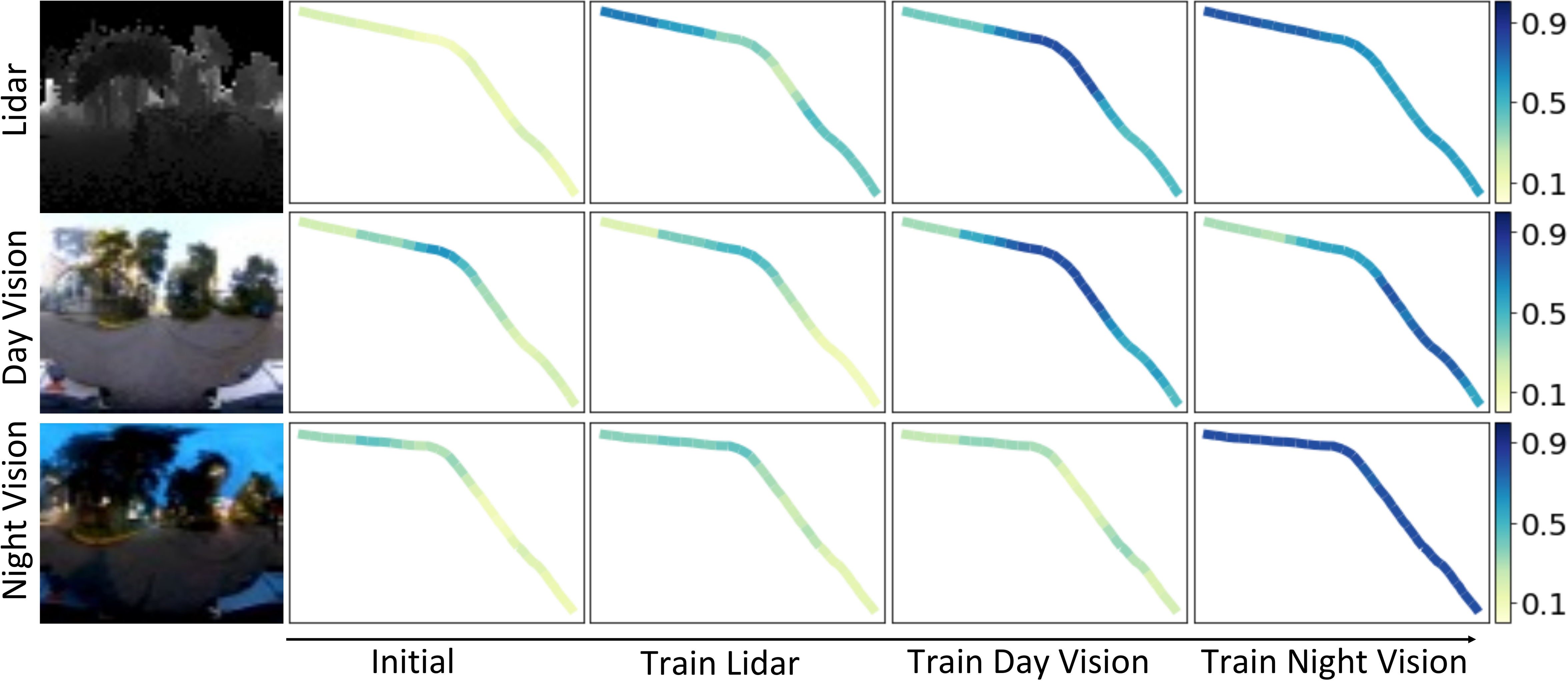}
    \label{fig:exp_cmu_confidence}
    }
    \vspace{-5pt}
    \caption{\textbf{Confidence map over training}. (a) City dataset. Sampled trajectories and confidence maps for area 1, area 2, area 3. (b) Campus dataset. Sampled trajectories and confidence maps for Lidar, day-time visual, and night-time visual observations. }
\end{figure*}

To illustrate the incremental learning property, we evaluate the BioSLAM with a confidence map during training. The confidence score of a place is defined by the cosine similarity between the corresponding reference $O_k$ and query $Q_k$ features, with  $\textit{confidence}(k)=\cos\left(\mathcal{F}(O_k), \mathcal{F}(Q_k)\right)$.  A higher $\textit{confidence}(k)$ denotes the robust feature representation of place $Q_k$, because the feature representations of query and reference in the same place are similar. The confidence map is measured by calculating the confidence score for all observations and visualizing the confidence scores on real trajectories.

The confidence map of BioSLAM over training on City dataset is shown in \cref{fig:exp_pitt_confidence}.
The left column represents the sampled trajectories from area 1, area 2, and area 3. The right three columns represent the confidence map of the corresponding trajectory  (from left to right) after incrementally training on area 1, area 2, and area 3, respectively.
After training on area 1, the confidence map of all areas becomes better. Then training on area 2 and area 3, the confidence map of area 1 is almost non-decayed. 
That means training BioSLAM on a trajectory also helps to improve the performance of others. 

The confidence map of BioSLAM over training on Campus dataset is shown in \cref{fig:exp_cmu_confidence}. The left column represents the sampled observations from Lidar, day-time visual, and night-time visual inputs of the same trajectory. The right three columns represent the confidence map of the corresponding domains after incrementally training on Lidar, day-time visual, and night-time visual inputs of the trajectory.
After training on the Lidar domain, the confidence map of the Lidar domain increases. 
Then incrementally training the model on the day-time and night-time visual domain, the confidence map of the corresponding domain increases, and other domains almost do not change. 
That means BioSLAM can remember past domains when learning from other domains. 

\begin{table}[ht]
    \centering
    \caption{Comparison of GPU memory (Megabyte) of different methods. 
    \label{table:time_efficiency}}
\begin{tabular}{c|cccc}
\hline
Method          & NetVLAD & SI   & GR   & BioSLAM \\ \hline
GPU Memory (MB) & 1261    & 1265 & 1695 & 1695    \\ \hline
\end{tabular}
\end{table}

    \subsection{Run-time Analysis}
\label{sec:run-time}

In this section, we introduce memory and time usage for lifelong learning.
For both methods, we evaluate an Ubuntu 18.04 system, using an Nvidia RTX 2080 Ti (12 GB) graphics processing unit (GPU), Intel Core i9-7900x processors, and $64$ gigabyte (GB) memory.
\Cref{table:time_efficiency} shows the total memory usage on City datasets.
The GPU memory usages of BioSLAM are acceptable under the current embedded system structure.

\Cref{fig:time_analsis} shows the time usage of the BioSLAM lifelong learning procedure on the City dataset when incrementally feeding new trajectory segments. The distance for each segment is about 2km and is composed of about 200 observation frames. 
The data inference procedure takes $<1$s for each trajectory segment, which is efficient in real-world inference.
The average time for memory consolidation and forgetting is around $40$s for each trajectory, which is fast enough to analyze the newly captured memory traces, and update the memory system.
And the memory replay takes $2$s to generate replayed samples for place recognition training.
Finally, place recognition optimization takes $40$s to optimize a $2$km new trajectory in one epoch. 
For multi-epoch training, BioSLAM runs inference, replay, and optimization multi-times, but only runs memory consolidation once. 
In general, training a trajectory segment about 50 times can get convergence results. 
So the total learning time for a 2km trajectory segment is $40{\rm s}+ (1+2+40)*50{\rm s}=2190{\rm s}\approx 36 {\rm min}$. 
Given that the distance between neighbor keyframes is $10$m, in this case, BioSLAM can learn $100$m new areas in around 1.8 minutes.

We can note the most important property, the time usage in the above memory operations will not be affected by scale differences in either spatial or temporal.
This is mainly benefited by our memory forgetting mechanism, which can maintain the searching space of $M_S, M_d$ and keep up-to-date memory traces to balance the localization.
The above properties indicate that BioSLAM can be applied to low-cost robotic systems for long-term place recognition tasks on NVIDIA embedded systems.    

\begin{figure}[t]
    \centering
    \includegraphics[width=\linewidth]{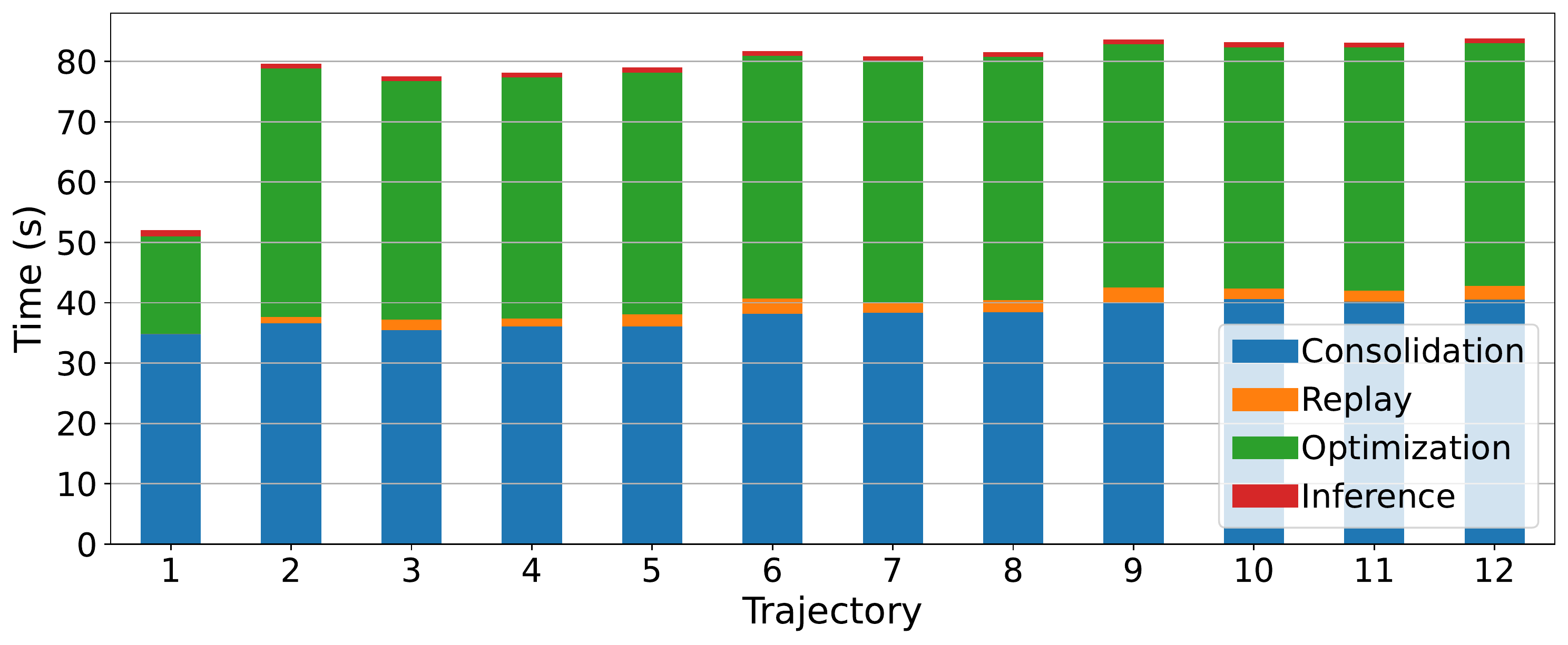}
    \color{red}
    \caption{\textbf{Time usage (inference, optimization, replay, consolidation time) of the BioSLAM lifelong learning on City dataset when incrementally learning new trajectories.}}
    \label{fig:time_analsis}
\end{figure}
    
    \section{Discussion \& Limitations}
\label{sec:discussion}
    BioSLAM can provide robust lifelong learning ability for long-term and large-scale place recognition tasks.
    For long-term localization tasks, BioSLAM can pre-store the long-lasting memory traces in the static memory $M_S$ and retrieval generative memories from dynamic memory $M_D$, which maintains the recognition ability for diverse conditions.
    The above dual-memory mechanism can enable efficient place feature learning for new types of observations and maintain the lifelong memorization ability for old knowledge.
    We can also notice that the model benefits from solving place recognition from multiple domains in lifelong learning. In the evaluation of the Campus dataset, for the BioSLAM method,  the knowledge of one domain can help better and faster understand other domains. 
    
    Finally, the essential property of our BioSLAM framework is its extensibility. \textit{We can develop a similar lifelong learning framework for other perception, navigation, or reinforcement learning tasks.}
    Let's recall the network structures as shown in \cref{fig:network}, the functional modules that related to the place recognition task is mainly the place descriptor extraction $\mathcal{F}_\theta$ and the relative external reward $\mathcal{R}_{ex}$ in the behavior configurator.
    For other tasks (such as 3D segmentation, local navigation, etc.), one can replace the place descriptor extraction network with a task-relative representation network, and objective reward, then not need to replace the entire blocks in the lifelong memory system.
    Also, another potential option is to develop a parallel hybrid lifelong learning system for multiple tasks since the encoder module $\mathcal{E}$ can be shared.

    In general, BioSLAM provides a memory system for lifelong place recognition.
    Robots can develop more general navigation and decision-making approaches based on the incrementally updated localization ability, given that most current systems can only work in short-term and local-scale environments.
    On the other hand, BioSLAM also provides a new option for other lifelong learning tasks.
    \section{Conclusion}
\label{sec:conclusion}
The real-world robots will encounter diverse environmental changes under long-term autonomy.
In the place recognition task, the robots continuously observe new scenarios, which are unbounded under variant conditions.
To alleviate the above problem, we proposed BioSLAM, a lifelong place recognition method in this work.
BioSLAM combines a general place learning (GPL) system and a bio-inspired lifelong memory (BiLM) system.
The GPL system utilizes a viewpoint-invariant place descriptor and a generative replay module to achieve the `memory encoding' and `memory replay' for continual place feature learning.
The BiLM system provides a dual-memory mechanism, controlled by a behavior configurator to guide the `memory consolidation', `memory forgetting', and `memory replay' to enhance the memorization of long-term traces.
We investigate the large-scale and long-term place recognition ability in the experiments with city-scale 3D point-cloud maps and campus-scale visual-LiDAR hybrid inputs. Both results show that BioSLAM can significantly balance the place learning ability for new observations and maintain the memorization ability for history observations.

In practice, our method can be applied to the low-cost mobile robots with the current embedded devices, with a lightweight memory system without saving massive streaming datasets.
Another interesting direction for future work is to enable memory sharing between client agents to the cloud server; in this case, the server can be synced with data from all kinds of scenarios by variant robots to update a more general place recognition.
Finally, the BioSLAM system can be utilized in other perception tasks via modifying objective functions in the behavior configurator by requirements.

\section{Acknowledgment}
This research is supported by grants from NVIDIA and utilized NVIDIA SDKs (CUDA Toolkit, TensorRT, and Omniverse).
This research is supported by the ARL grant NO.W911QX20D0008 and partially supported by the National Science Foundation (NSF) under Grant No. 2144489. 
Any opinions, findings, conclusions, or recommendations expressed in this material are those of the author(s) and do not necessarily reflect the views of ARL and NSF.
    \bibliographystyle{IEEEtran}
    \bibliography{bible}

\begin{thebibliography}{10}
\providecommand{\url}[1]{#1}
\csname url@samestyle\endcsname
\providecommand{\newblock}{\relax}
\providecommand{\bibinfo}[2]{#2}
\providecommand{\BIBentrySTDinterwordspacing}{\spaceskip=0pt\relax}
\providecommand{\BIBentryALTinterwordstretchfactor}{4}
\providecommand{\BIBentryALTinterwordspacing}{\spaceskip=\fontdimen2\font plus
\BIBentryALTinterwordstretchfactor\fontdimen3\font minus
  \fontdimen4\font\relax}
\providecommand{\BIBforeignlanguage}[2]{{%
\expandafter\ifx\csname l@#1\endcsname\relax
\typeout{** WARNING: IEEEtran.bst: No hyphenation pattern has been}%
\typeout{** loaded for the language `#1'. Using the pattern for}%
\typeout{** the default language instead.}%
\else
\language=\csname l@#1\endcsname
\fi
#2}}
\providecommand{\BIBdecl}{\relax}
\BIBdecl

\bibitem{Survey:SLAM}
C.~Cadena, L.~Carlone, H.~Carrillo, Y.~Latif, D.~Scaramuzza, J.~Neira, I.~Reid,
  and J.~J. Leonard, ``Past, present, and future of simultaneous localization
  and mapping: Toward the robust-perception age,'' \emph{IEEE Transactions on
  robotics}, vol.~32, no.~6, pp. 1309--1332, 2016.

\bibitem{detone2018superpoint}
D.~DeTone, T.~Malisiewicz, and A.~Rabinovich, ``Superpoint: Self-supervised
  interest point detection and description,'' in \emph{Proceedings of the IEEE
  conference on computer vision and pattern recognition workshops}, 2018, pp.
  224--236.

\bibitem{latif2018addressing}
Y.~Latif, R.~Garg, M.~Milford, and I.~Reid, ``Addressing challenging place
  recognition tasks using generative adversarial networks,'' in \emph{2018 IEEE
  International Conference on Robotics and Automation (ICRA)}, 2018, pp.
  2349--2355.

\bibitem{i3dloc}
P.~Yin, L.~Xu, J.~Zhang, H.~Choset, and S.~Scherer, ``i3dloc: Image-to-range
  cross-domain localization robust to inconsistent environmental conditions,''
  in \emph{Proceedings of Robotics: Science and Systems (RSS '21)}.\hskip 1em
  plus 0.5em minus 0.4em\relax Robotics: Science and Systems 2021, 2021.

\bibitem{Hip:Place_cell}
E.~I. Moser, E.~Kropff, and M.-B. Moser, ``Place cells, grid cells, and the
  brain's spatial representation system,'' \emph{Annu. Rev. Neurosci.},
  vol.~31, pp. 69--89, 2008.

\bibitem{frontal_lobe}
J.~Stretton and P.~Thompson, ``Frontal lobe function in temporal lobe
  epilepsy,'' \emph{Epilepsy research}, vol.~98, no.~1, pp. 1--13, 2012.

\bibitem{hippo_memory}
G.~Berdugo-Vega and J.~Graeff, ``Inquiring the librarian about the location of
  memory,'' \emph{Cognitive Neuroscience}, vol.~0, no.~0, pp. 1--3, 2022.

\bibitem{klinzing2019mechanisms}
J.~G. Klinzing, N.~Niethard, and J.~Born, ``Mechanisms of systems memory
  consolidation during sleep,'' \emph{Nature neuroscience}, vol.~22, no.~10,
  pp. 1598--1610, 2019.

\bibitem{VPR_Bench}
M.~Zaffar, S.~Garg, M.~Milford, J.~Kooij, D.~Flynn, K.~McDonald-Maier, and
  S.~Ehsan, ``{VPR}-bench: An open-source visual place recognition evaluation
  framework with quantifiable viewpoint and appearance change,''
  \emph{International Journal of Computer Vision.}, vol. 129, no.~7, pp.
  2136--2174, 2021.

\bibitem{VPR:SURVEY}
S.~Lowry, N.~Sünderhauf, P.~Newman, J.~J. Leonard, D.~Cox, P.~Corke, and M.~J.
  Milford, ``Visual place recognition: A survey,'' \emph{IEEE Transactions on
  Robotics}, vol.~32, no.~1, pp. 1--19, 2016.

\bibitem{deep_pr_survey}
T.~Barros, R.~Pereira, L.~Garrote, C.~Premebida, and U.~J. Nunes, ``Place
  recognition survey: An update on deep learning approaches,'' \emph{arXiv
  preprint arXiv:2106.10458}, 2021.

\bibitem{deep_vpr_survey}
X.~Zhang, L.~Wang, and Y.~Su, ``Visual place recognition: A survey from deep
  learning perspective,'' \emph{Pattern Recognition}, vol. 113, p. 107760,
  2021.

\bibitem{lifelong_survey}
M.~Delange, R.~Aljundi, M.~Masana, S.~Parisot, X.~Jia, A.~Leonardis,
  G.~Slabaugh, and T.~Tuytelaars, ``A continual learning survey: Defying
  forgetting in classification tasks,'' \emph{IEEE Transactions on Pattern
  Analysis and Machine Intelligence}, 2021.

\bibitem{CLRobot}
T.~Lesort, V.~Lomonaco, A.~Stoian, D.~Maltoni, D.~Filliat, and N.~D.
  Rodr{\'i}guez, ``Continual learning for robotics: Definition, framework,
  learning strategies, opportunities and challenges,'' \emph{Information
  Fusion}, vol.~58, pp. 52--68, 2020.

\bibitem{FEATURE:SIFT}
K.~Mikolajczyk and C.~Schmid, ``A performance evaluation of local
  descriptors,'' \emph{IEEE Transactions on Pattern Analysis and Machine
  Intelligence.}, vol.~27, no.~10, pp. 1615--1630, 2005.

\bibitem{feature:orb}
E.~Rublee, V.~Rabaud, K.~Konolige, and G.~Bradski, ``{ORB}: An efficient
  alternative to {SIFT} or {SURF},'' in \emph{2011 International Conference on
  Computer Vision.}, 2011, pp. 2564--2571.

\bibitem{fabmap}
M.~Cummins and P.~Newman, ``Fab-map: Probabilistic localization and mapping in
  the space of appearance,'' \emph{The International Journal of Robotics
  Research}, vol.~27, no.~6, pp. 647--665, 2008.

\bibitem{ibow_lcd}
E.~Garcia-Fidalgo and A.~Ortiz, ``ibow-lcd: An appearance-based loop-closure
  detection approach using incremental bags of binary words,'' \emph{IEEE
  Robotics and Automation Letters}, vol.~3, no.~4, pp. 3051--3057, 2018.

\bibitem{bow_fast}
S.~An, H.~Zhu, D.~Wei, K.~A. Tsintotas, and A.~Gasteratos, ``Fast and
  incremental loop closure detection with deep features and proximity graphs,''
  \emph{Journal Of Field Robotics}, vol.~39, no.~4, pp. 473--493, 2022.

\bibitem{Feature:VGG}
K.~Simonyan and A.~Zisserman, ``Very deep convolutional networks for
  large-scale image recognition,'' \emph{arXiv preprint arXiv:1409.1556}, 2014.

\bibitem{resnet}
K.~He, X.~Zhang, S.~Ren, and J.~Sun, ``Identity mappings in deep residual
  networks,'' in \emph{European conference on computer vision}.\hskip 1em plus
  0.5em minus 0.4em\relax Springer, 2016, pp. 630--645.

\bibitem{transformer}
A.~Vaswani, N.~Shazeer, N.~Parmar, J.~Uszkoreit, L.~Jones, A.~N. Gomez,
  {\L}.~Kaiser, and I.~Polosukhin, ``Attention is all you need,''
  \emph{Advances in neural information processing systems}, vol.~30, 2017.

\bibitem{NetVLAD}
R.~Arandjelovic, P.~Gronat, A.~Torii, T.~Pajdla, and J.~Sivic, ``Netvlad: Cnn
  architecture for weakly supervised place recognition,'' in \emph{Proceedings
  of the IEEE conference on computer vision and pattern recognition.}, 2016,
  pp. 5297--5307.

\bibitem{VLAD}
R.~Arandjelovic and A.~Zisserman, ``All about vlad,'' in \emph{2013 IEEE
  Conference on Computer Vision and Pattern Recognition}, 2013, pp. 1578--1585.

\bibitem{regionVLAD}
A.~Khaliq, S.~Ehsan, Z.~Chen, M.~Milford, and K.~McDonald-Maier, ``A holistic
  visual place recognition approach using lightweight cnns for significant
  viewpoint and appearance changes,'' \emph{IEEE Transactions on Robotics},
  vol.~36, no.~2, pp. 561--569, 2020.

\bibitem{2019conditionVLAD}
J.~M. Facil, D.~Olid, L.~Montesano, and J.~Civera, ``Condition-invariant
  multi-view place recognition,'' \emph{arXiv preprint arXiv:1902.09516}, 2019.

\bibitem{hui2021pyramid}
L.~Hui, H.~Yang, M.~Cheng, J.~Xie, and J.~Yang, ``Pyramid point cloud
  transformer for large-scale place recognition,'' in \emph{Proceedings of the
  IEEE/CVF International Conference on Computer Vision}, 2021, pp. 6098--6107.

\bibitem{m2dp}
L.~He, X.~Wang, and H.~Zhang, ``M2dp: A novel 3d point cloud descriptor and its
  application in loop closure detection,'' in \emph{2016 IEEE/RSJ International
  Conference on Intelligent Robots and Systems (IROS)}, 2016, pp. 231--237.

\bibitem{scancontext}
G.~Kim and A.~Kim, ``Scan context: Egocentric spatial descriptor for place
  recognition within 3d point cloud map,'' in \emph{2018 IEEE/RSJ International
  Conference on Intelligent Robots and Systems (IROS)}.\hskip 1em plus 0.5em
  minus 0.4em\relax IEEE, 2018, pp. 4802--4809.

\bibitem{PR:pointnetvlad}
M.~A. Uy and G.~H. Lee, ``Pointnetvlad: Deep point cloud based retrieval for
  large-scale place recognition,'' in \emph{2018 {IEEE} Conference on Computer
  Vision and Pattern Recognition,}, 2018, pp. 4470--4479.

\bibitem{pr:lpdnet}
Z.~Liu, S.~Zhou, C.~Suo, P.~Yin, W.~Chen, H.~Wang, H.~Li, and Y.~Liu,
  ``Lpd-net: 3d point cloud learning for large-scale place recognition and
  environment analysis,'' in \emph{2019 {IEEE/CVF} International Conference on
  Computer Vision, {ICCV}}, 2019, pp. 2831--2840.

\bibitem{pcan}
W.~Zhang and C.~Xiao, ``Pcan: 3d attention map learning using contextual
  information for point cloud based retrieval,'' in \emph{Proceedings of the
  IEEE/CVF Conference on Computer Vision and Pattern Recognition}, 2019, pp.
  12\,436--12\,445.

\bibitem{PR:overlapnet}
X.~Chen, T.~L\"abe, A.~Milioto, T.~R\"ohling, O.~Vysotska, A.~Haag, J.~Behley,
  and C.~Stachniss, ``{OverlapNet: Loop Closing for LiDAR-based SLAM},'' in
  \emph{Proceedings of Robotics: Science and Systems (RSS)}, 2020.

\bibitem{yin2021fusionvlad}
P.~Yin, L.~Xu, J.~Zhang, and H.~Choset, ``Fusionvlad: A multi-view deep fusion
  networks for viewpoint-free 3d place recognition,'' \emph{IEEE Robotics and
  Automation Letters}, vol.~6, no.~2, pp. 2304--2310, 2021.

\bibitem{yin2021fast}
P.~Yin, F.~Wang, A.~Egorov, J.~Hou, Z.~Jia, and J.~Han, ``Fast
  sequence-matching enhanced viewpoint-invariant 3-d place recognition,''
  \emph{IEEE Transactions on Industrial Electronics}, vol.~69, no.~2, pp.
  2127--2135, 2021.

\bibitem{yin2020seqspherevlad}
P.~Yin, F.~Wang, A.~Egorov, J.~Hou, J.~Zhang, and H.~Choset, ``Seqspherevlad:
  Sequence matching enhanced orientation-invariant place recognition,'' in
  \emph{2020 IEEE/RSJ International Conference on Intelligent Robots and
  Systems (IROS)}, 2020, pp. 5024--5029.

\bibitem{milford2012seqslam}
M.~J. Milford and G.~F. Wyeth, ``Seqslam: Visual route-based navigation for
  sunny summer days and stormy winter nights,'' in \emph{2012 IEEE
  international conference on robotics and automation}, 2012, pp. 1643--1649.

\bibitem{lwf}
Z.~Li and D.~Hoiem, ``Learning without forgetting,'' \emph{IEEE transactions on
  pattern analysis and machine intelligence}, vol.~40, no.~12, pp. 2935--2947,
  2017.

\bibitem{packnet}
A.~Mallya and S.~Lazebnik, ``Packnet: Adding multiple tasks to a single network
  by iterative pruning,'' in \emph{Proceedings of the IEEE conference on
  Computer Vision and Pattern Recognition}, 2018, pp. 7765--7773.

\bibitem{EWC}
J.~Kirkpatrick, R.~Pascanu, N.~Rabinowitz, J.~Veness, G.~Desjardins, A.~A.
  Rusu, K.~Milan, J.~Quan, T.~Ramalho, A.~Grabska-Barwinska \emph{et~al.},
  ``Overcoming catastrophic forgetting in neural networks,'' \emph{Proceedings
  of the national academy of sciences}, vol. 114, no.~13, pp. 3521--3526, 2017.

\bibitem{SI}
F.~Zenke, B.~Poole, and S.~Ganguli, ``Continual learning through synaptic
  intelligence,'' in \emph{International Conference on Machine Learning}.\hskip
  1em plus 0.5em minus 0.4em\relax PMLR, 2017, pp. 3987--3995.

\bibitem{icarl}
S.-A. Rebuffi, A.~Kolesnikov, G.~Sperl, and C.~H. Lampert, ``icarl: Incremental
  classifier and representation learning,'' in \emph{Proceedings of the IEEE
  conference on Computer Vision and Pattern Recognition}, 2017, pp. 2001--2010.

\bibitem{gem}
D.~Lopez-Paz and M.~Ranzato, ``Gradient episodic memory for continual
  learning,'' \emph{Advances in neural information processing systems},
  vol.~30, 2017.

\bibitem{GP}
G.~M. Van~de Ven and A.~S. Tolias, ``Generative replay with feedback
  connections as a general strategy for continual learning,'' \emph{arXiv
  preprint arXiv:1809.10635}, 2018.

\bibitem{choi2021dual}
Y.~Choi, M.~El-Khamy, and J.~Lee, ``Dual-teacher class-incremental learning
  with data-free generative replay,'' in \emph{Proceedings of the IEEE/CVF
  Conference on Computer Vision and Pattern Recognition}, 2021, pp. 3543--3552.

\bibitem{stickgold2005sleep}
R.~Stickgold, ``Sleep-dependent memory consolidation,'' \emph{Nature}, vol.
  437, no. 7063, pp. 1272--1278, 2005.

\bibitem{Sphere:SO3_learning}
C.~Esteves, C.~Allen-Blanchette, A.~Makadia, and K.~Daniilidis, ``Learning so
  (3) equivariant representations with spherical cnns,'' in \emph{Proceedings
  of the European Conference on Computer Vision (ECCV).}, 2018, pp. 52--68.

\bibitem{cohen2018spherical}
T.~S. Cohen, M.~Geiger, J.~K{\"{o}}hler, and M.~Welling, ``Spherical cnns,'' in
  \emph{6th International Conference on Learning Representations, {ICLR}},
  2018.

\bibitem{cnn:gan}
I.~Goodfellow, J.~Pouget-Abadie, M.~Mirza, B.~Xu, D.~Warde-Farley, S.~Ozair,
  A.~Courville, and Y.~Bengio, ``Generative adversarial nets,'' \emph{Advances
  in neural information processing systems.}, vol.~27, 2014.

\bibitem{Shin2017deepreplay}
H.~Shin, J.~K. Lee, J.~Kim, and J.~Kim, ``Continual learning with deep
  generative replay,'' in \emph{ADVANCES IN NEURAL INFORMATION PROCESSING
  SYSTEMS 30 (NIPS 2017)}, ser. Advances in Neural Information Processing
  Systems, I.~Guyon, U.~Luxburg, S.~Bengio, H.~Wallach, R.~Fergus,
  S.~Vishwanathan, and R.~Garnett, Eds., vol.~30, 2017.

\bibitem{vandeven2020brain}
G.~M. van~de Ven, H.~T. Siegelmann, and A.~S. Tolias, ``Brain-inspired replay
  for continual learning with artificial neural networks,'' \emph{Nature
  Communications}, vol.~11, p. 4069, 2020.

\bibitem{krueger2009flexible}
K.~A. Krueger and P.~Dayan, ``Flexible shaping: How learning in small steps
  helps,'' \emph{Cognition}, vol. 110, no.~3, pp. 380--394, 2009.

\bibitem{bengio2009curriculum}
Y.~Bengio, J.~Louradour, R.~Collobert, and J.~Weston, ``Curriculum learning,''
  in \emph{Proceedings of the 26th annual international conference on machine
  learning}, 2009, pp. 41--48.

\bibitem{yin2022automerge}
P.~Yin, L.~Haowen, S.~Zhao, R.~Fu, C.~Ivan, R.~Ge, I.~Cisneros, R.~Fu,
  J.~Zhang, H.~Choset, and S.~Scherer, ``Automerge: A framework for map
  assembling and smoothing in city-scale environments,'' \emph{arXiv preprint
  arXiv:2205.19737}, 2022.

\bibitem{byrne2017learning}
J.~Byrne, \emph{Learning and memory: a comprehensive reference}.\hskip 1em plus
  0.5em minus 0.4em\relax Academic Press, 2017.

\bibitem{berman2009search}
M.~G. Berman, J.~Jonides, and R.~L. Lewis, ``In search of decay in verbal
  short-term memory.'' \emph{Journal of Experimental Psychology: Learning,
  Memory, and Cognition}, vol.~35, no.~2, p. 317, 2009.

\bibitem{LOAM:zhang2014loam}
J.~Zhang and S.~Singh, ``Loam: Lidar odometry and mapping in real-time.'' in
  \emph{Robotics Science and Systems}, vol.~2, no.~9.\hskip 1em plus 0.5em
  minus 0.4em\relax Berkeley, CA, 2014, pp. 1--9.

\bibitem{segal2009generalized}
A.~Segal, D.~Haehnel, and S.~Thrun, ``Generalized-icp.'' in \emph{Robotics:
  science and systems}, vol.~2, no.~4, 2009, p. 435.

\bibitem{VPR:DBOW2}
D.~G\'alvez-L\'opez and J.~D. Tard\'os, ``Bags of binary words for fast place
  recognition in image sequences,'' \emph{IEEE Transactions on Robotics.},
  vol.~28, no.~5, pp. 1188--1197, October 2012.

\bibitem{VPR:CoHOG}
M.~Zaffar, S.~Ehsan, M.~Milford, and K.~McDonald-Maier, ``Cohog: A
  light-weight, compute-efficient, and training-free visual place recognition
  technique for changing environments,'' \emph{IEEE Robotics and Automation
  Letters.}, vol.~5, no.~2, pp. 1835--1842, 2020.

\bibitem{VPR:RegionVLAD}
A.~Khaliq, S.~Ehsan, Z.~Chen, M.~Milford, and K.~McDonald-Maier, ``A holistic
  visual place recognition approach using lightweight cnns for significant
  viewpoint and appearance changes,'' \emph{IEEE Transactions on Robotics.},
  vol.~36, no.~2, pp. 561--569, 2020.

\bibitem{generative_play}
H.~Shin, J.~K. Lee, J.~Kim, and J.~Kim, ``Continual learning with deep
  generative replay,'' \emph{Advances in neural information processing
  systems}, vol.~30, 2017.

\bibitem{Infants_ability}
D.~A. Baldwin, E.~M. Markman, and R.~L. Melartin, ``Infants' ability to draw
  inferences about nonobvious object properties: Evidence from exploratory
  play,'' \emph{Child Development}, vol.~64, no.~3, pp. 711--728, 1993.

\bibitem{bornstein2010development}
M.~H. Bornstein and M.~E. Arterberry, ``The development of object
  categorization in young children: hierarchical inclusiveness, age, perceptual
  attribute, and group versus individual analyses.'' \emph{Developmental
  psychology}, vol.~46, no.~2, p. 350, 2010.

\end{thebibliography}
    \vspace{-1cm}
\begin{IEEEbiography}[{\includegraphics[width=1in,height=1.25in,clip,keepaspectratio]{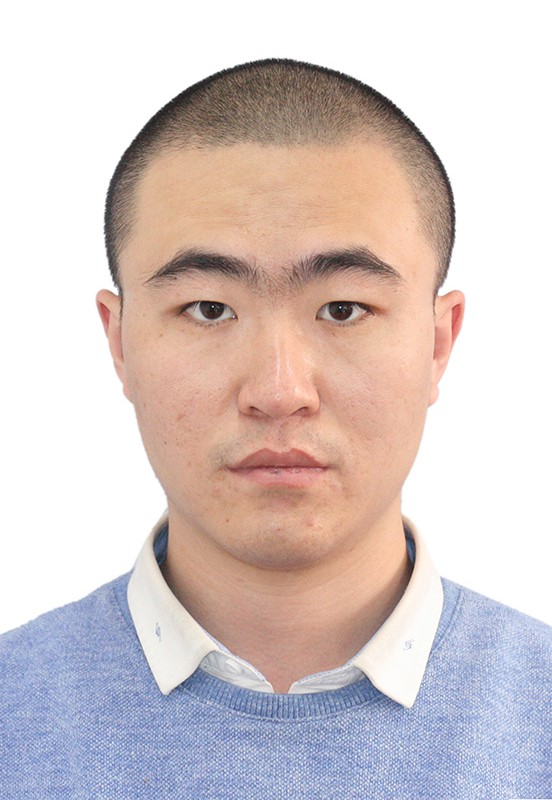}}]
    {Peng Yin} received his Bachelor's degree from Harbin Institute of Technology, Harbin, China, in 2013, and his Ph.D. degree from the University of Chinese Academy of Sciences, Beijing, in 2018.
    He is research Post-doctoral with the Department of the Robotics Institute, Carnegie Mellon University, Pittsburgh, USA.
    His research interests include LiDAR SLAM, Place Recognition, 3D Perception, and Reinforcement Learning. Dr. Yin has served as a Reviewer for several IEEE Conferences ICRA, IROS, ACC, and RSS.
\end{IEEEbiography}

\begin{IEEEbiography}[{\includegraphics[width=1in,height=1.25in,clip,keepaspectratio]{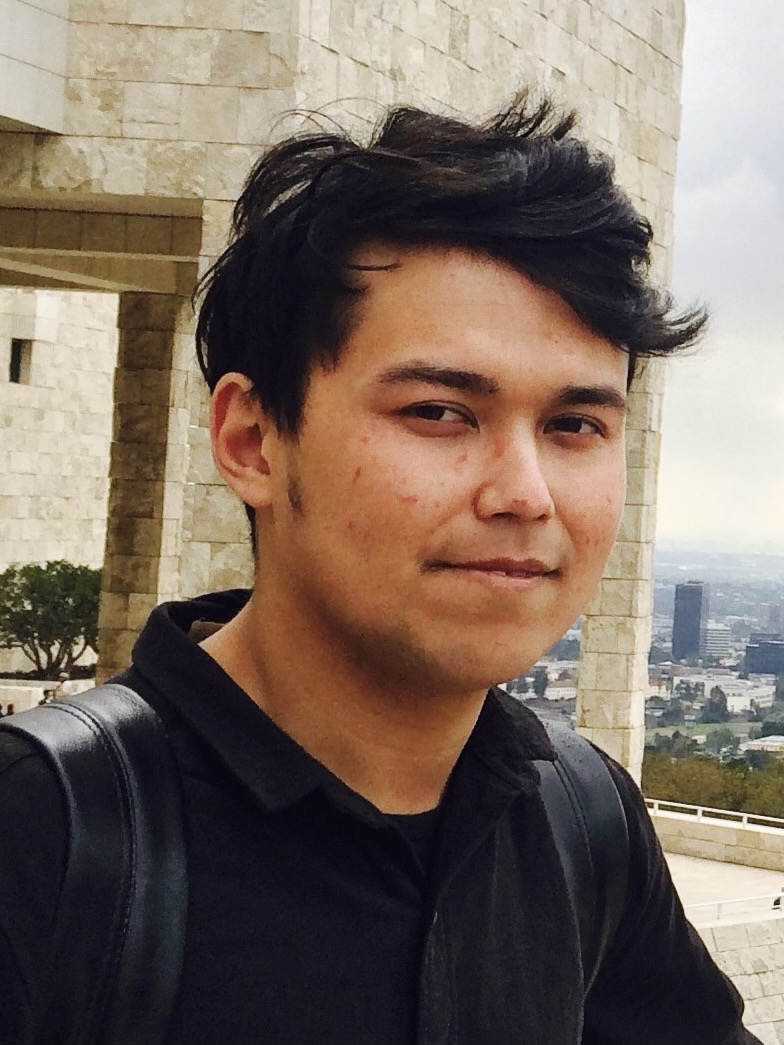}}]
    {Abulikemu Abuduweili} received his B.S. in Electrical Engineering from Peking University, China, in 2017, and his M.S. in Electrical Engineering from Peking University, China, in 2020.  
    He is currently a PhD student at Electrical Engineering and Robotics at Carnegie Mellon University, USA.
    His research interests include Deep Learning, Robotics, and Computer Vision.
\end{IEEEbiography}

\begin{IEEEbiography}[{\includegraphics[width=1in,height=1.25in,clip,keepaspectratio]{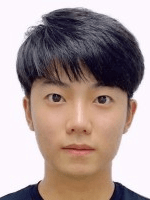}}]
    {Shiqi Zhao} received his Bachelor's degree from Dalian University of Technology, Dalian, China, in 2018, and his Master's degree from the University of California San Diego, U.S., in 2020.
    He works as an intern at the Robotics Institute at Carnegie Mellon University.
    His research interests include Place Recognition, 3D Perception, and Deep Learning.
\end{IEEEbiography}

\begin{IEEEbiography}[{\includegraphics[width=1in,height=1.25in,clip,keepaspectratio]{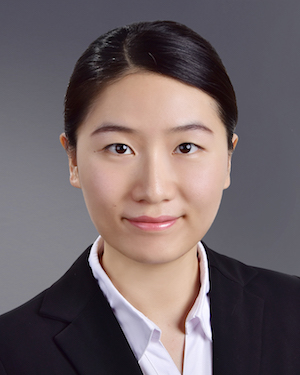}}]
    {Changliu Liu} received the B.S. degree in mechanical engineering and the B.S. degree in economics from Tsinghua University, China, in 2012, the M.S. degree in mechanical engineering, and the M.A. degree in mathematics from the University of California, Berkeley, U.S.A., in 2014 and 2016 respectively, and the Ph.D. degree in mechanical engineering from the University of California, Berkeley, U.S.A., in 2017. She is an assistant professor at the Robotics Institute at Carnegie Mellon University. 
    Her research interests lie in designing and verifying intelligent systems with applications to manufacturing and transportation. She received NSF Career Award, Amazon Research Award, and Ford URP Award.

\end{IEEEbiography}

\begin{IEEEbiography}[{\includegraphics[width=1in,height=1.25in,clip,keepaspectratio]{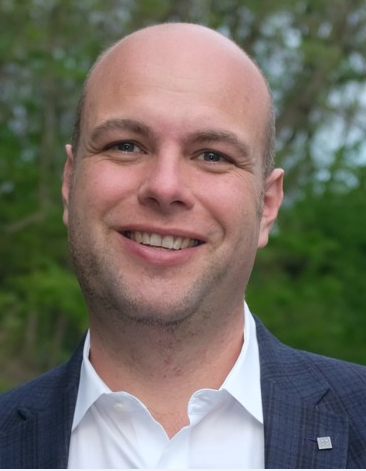}}]
    {Sebastian Scherer} received his B.S. in Computer Science, M.S., and Ph.D. in Robotics from CMU in 2004, 2007, and 2010. 
    Sebastian Scherer is an Associate Research Professor at the Robotics Institute at Carnegie Mellon University. His research focuses on enabling autonomous unmanned rotorcraft to operate at low altitudes in cluttered environments. He is a Siebel scholar and a recipient of multiple paper awards and nominations, including AIAA@Infotech 2010 and FSR 2013. His research has been covered by the national and internal press, including IEEE Spectrum, the New Scientist, Wired, der Spiegel, and the WSJ. 
\end{IEEEbiography}
\endgroup

\end{document}